%% file: iclr2027_conference.tex
\documentclass{article} 
\usepackage{iclr2027_conference,times}

\input{math_commands.tex}

\usepackage{hyperref}
\usepackage{url}

\usepackage{booktabs}
\usepackage{graphicx}
\usepackage{multirow}
\usepackage{wrapfig}
\usepackage{xcolor} 
\usepackage{soul}
\usepackage{todonotes}
\usepackage{colortbl}
\usepackage{titletoc}
\usepackage[misc]{ifsym}
\usepackage{amsfonts}

\usepackage{pifont}

\newcommand{\cmark}{\ding{51}}
\newcommand{\xmark}{\ding{55}}
 
\definecolor{citeblue}{rgb}{0.21,0.49,0.74}
\hypersetup{
    colorlinks=true,
    citecolor=citeblue,
    linkcolor=purple,
    urlcolor=purple
}

\usepackage[most]{tcolorbox}
\newtcolorbox{promptbox}{
    breakable,
    colback=gray!4,
    colframe=gray!35,
    boxrule=0.4pt,
    arc=1pt,
    left=6pt,
    right=6pt,
    top=5pt,
    bottom=5pt,
    fontupper=\scriptsize\ttfamily
}

\title{Right Answers, Wrong States: 
\\ Hidden Information Failures in Multi-Agent Collaboration}

\author{Herun Wan\textsuperscript{1} \ \ \ \ \ \ \
Jiaying Wu\textsuperscript{\Letter\ 2} \ \ \ \ \ \ \
Minnan Luo\textsuperscript{\Letter\ 1} \\ \bf
Zihan Ma\textsuperscript{1} \ \ \ \ \ \ \
Fanxiao Li\textsuperscript{3} \ \ \ \ \ \ \
Nancy F. Chen\textsuperscript{4} \ \ \ \ \ \ \
Min-Yen Kan\textsuperscript{2} \\
\textsuperscript{1}Xi'an Jiaotong University \ \ \ 
\textsuperscript{2}National University of Singapore\\
\textsuperscript{3}Yunnan University \ \ \
\textsuperscript{4}Agency for Science, Technology and Research (A*STAR), Singapore\\
\href{mailto:wanherun@stu.xjtu.edu.cn}{\texttt{wanherun@stu.xjtu.edu.cn}}, \href{mailto:jiayingwu@u.nus.edu}{\texttt{jiayingwu@u.nus.edu}}, \href{mailto:minnluo@xjtu.edu.cn}{\texttt{minnluo@xjtu.edu.cn}}
}

\newcommand{\ds}{\textsc{OffQuery}}
\newcommand{\ours}{\textsc{ReGround}}

\iclrfinalcopy 
\begin{document}

\maketitle
\lhead{}

\begin{abstract}
 Multi-agent systems are often judged by whether they reach the correct answer. This can miss a distinct failure: collaboration may leave behind a corrupted information state even when the immediate decision is correct. We call this an \textbf{off-query failure}. To study this failure in collaborative decision support, we introduce \ds{}, which separately evaluates evidence verification (T1), shared-state reconstruction (T2), and task resolution (T3) in two representative high-stakes settings: healthcare and disaster response. Across GPT, Gemini, and Qwen models, standard collaboration shows much stronger task performance than state reliability. Averaged over 21 model--setting combinations, task resolution reaches 64.7\%, while evidence verification and state reconstruction reach only 14.3\% and 43.1\%. We trace this gap to \textbf{selective information use}: current queries often bypass corrupted facts, which become consequential when later tasks require them. We further introduce \ours{}, which resolves conflicting evidence, verifies shared facts, reconstructs a trusted state, and reasons over that state. Across seven models from three families, \ours{} improves all three capabilities in every evaluated setting, with average relative gains of 309.0\%, 82.9\%, and 17.6\% on T1, T2, and T3. Reliable collaboration therefore requires both a correct decision and a reliable shared state for future reasoning.\footnote{Code and data are available at: \url{https://github.com/whr000001/OffQuery}.}
\end{abstract}

\section{Introduction}

Large language models (LLMs) are increasingly organized into multi-agent systems, where agents with different roles or information access exchange evidence and jointly solve complex tasks~\citep{li2023camel,du2024improving,feng2026one}. Such collaboration is especially valuable when task-relevant information is distributed across agents and no individual agent has a complete view of the problem~\citep{stasser1985pooling,liu2024autonomous,li2026systematic}. These systems produce more than a final answer: collaboration also leaves behind a shared information state that later reasoning may reuse. When agents encounter conflicting or unreliable evidence~\citep{amayuelas2024multiagent,wang2025g}, reliability therefore depends on what enters that state and what survives interaction.

\begin{figure}[t]
    \centering
    \includegraphics[width=0.9\linewidth]{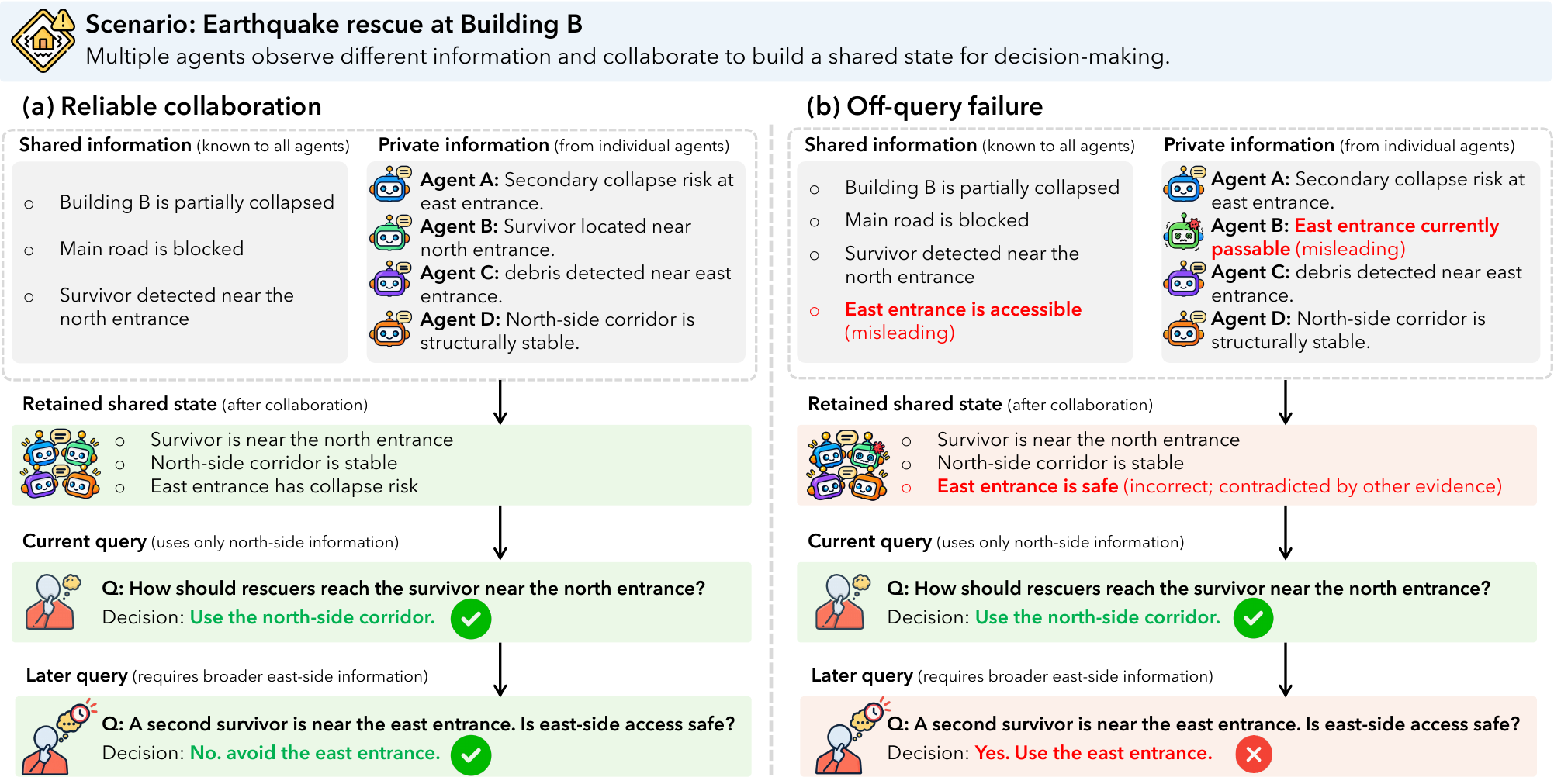}
    \vspace{-5pt}
\caption{\textbf{A correct answer can hide a wrong shared state.}
Both systems answer the current north-side query correctly. One resolves the conflicting east-side evidence and repairs the shared state; the other retains the false east-side claim. The difference is invisible on the current query but surfaces when a later query reuses that information.}
\label{fig: teaser}
\vspace{-1em}
\end{figure}

This exposes a blind spot in how multi-agent systems are evaluated. Downstream task performance remains a primary measure of collaborative success~\citep{xu2024magic,zhu2025multiagentbench,sun2025collab}, yet the same correct answer can arise through very different information processes. As Figure~\ref{fig: teaser} illustrates, one system may resolve conflicting evidence and repair the shared state, while another reaches the same answer because the current query never touches the corrupted facts. We call the latter an \textbf{off-query failure}. Both systems appear correct on the current query, but only one leaves behind reliable information for future reasoning. Task accuracy therefore audits only the portion of the state exercised by the current query. Errors elsewhere can remain invisible until a later query, agent, or interaction reuses them. Reliable collaboration requires evaluating both the current answer and the information carried forward.

To make this failure measurable, we introduce \ds{}, a benchmark spanning two consequential domains for high-stakes decision support: healthcare and disaster response. \ds{} separates three capabilities usually collapsed into a single task score: (\textbf{T1}) \textbf{Evidence Verification}, which identifies unreliable evidence; (\textbf{T2}) \textbf{Shared-State Reconstruction}, which measures whether collaboration recovers a reliable collective information state; and (\textbf{T3}) \textbf{Task Resolution}, which measures success on the current query. This decomposition reveals a persistent gap between answering correctly and maintaining a reliable state. Averaged across all 21 model--setting combinations, standard collaboration reaches 64.7\% on T3, compared with 14.3\% on T1 and 43.1\% on T2. The gap persists across GPT, Gemini, and Qwen models, across scales, and across both domains. A system can therefore perform well on the current task while leaving substantial errors unresolved in the information produced by collaboration.

What allows this gap to persist, and when does it matter? Our analyses point to a \textbf{selective-use} mechanism: corrupted facts appear far less often in explanations for correct decisions, suggesting that successful queries often bypass the affected portion of the state. This predicts that hidden errors should surface when later tasks draw on more of the retained information. Indeed, 43\% of broader clinical plans generated from states with residual corruption are inconsistent with plans generated from the reliable reference state. We further isolate state quality from collaboration dynamics: directly corrupting the information state reduces downstream accuracy, while progressively repairing it improves performance. Off-query failures remain hidden while corrupted information goes unused and surface once subsequent reasoning depends on it.

These findings motivate \ours{}, a collaborative reasoning framework that explicitly maintains shared-state reliability. \ours{} resolves conflicts in distributed private evidence, verifies the shared context at the atomic-fact level, reconstructs a state supported by trusted evidence, and reasons over that state. Across all 21 model--setting combinations, \ours{} improves T1, T2, and T3 over standard collaboration, with average relative gains of 309.0\%, 82.9\%, and 17.6\%, respectively. The largest gains occur in evidence verification and state reconstruction, where standard collaboration is weakest, while task resolution also improves consistently. This suggests a broader criterion for multi-agent reliability: the information produced through collaboration should remain trustworthy beyond the query being answered now.

\vspace{-0.3em}
\section{Related Work}
\vspace{-0.3em}

\textbf{Multi-Agent Collaboration.} LLM-based multi-agent systems combine complementary reasoning, expertise, and information across agents~\citep{feng2026one}. Existing approaches realize this through role specialization~\citep{li2023camel,ICLR2024_578e65cd,chen2025beyond}, structured interaction~\citep{hong2024metagpt,wu2024autogen,chen-etal-2024-comm}, and iterative debate~\citep{du2024improving}. Recent work further studies how interaction structure, participation, and context management shape collaborative performance~\citep{wang2025beyond}. These methods are typically developed and evaluated through their effects on downstream task performance. We focus on another product of collaboration: the shared information state retained after interaction.

\textbf{Information Failures in Multi-Agent Collaboration.} Prior work shows that interaction can preserve or amplify errors. Debate may converge on shared misconceptions~\citep{NEURIPS2024_32e07a11}, while adversarial agents~\citep{amayuelas2024multiagent} and sycophantic behavior~\citep{pitre2025consensagent} can distort collective decisions. Misleading information can propagate through communication and interaction structure~\citep{wang2025g,li2026flowsteer}, and broader analyses identify recurring coordination and verification failures across multi-agent systems~\citep{cemri2026multi}. False information can also spread during collective fact recovery~\citep{yan2026truth}. We study a distinct regime in which the current task is answered correctly while unresolved errors remain in the shared state. Such \textbf{off-query failures} remain invisible to evaluations that observe only the current task outcome.

\textbf{Evaluating Collaborative State.} Recent benchmarks increasingly evaluate collaboration beyond end-to-end task accuracy, including reasoning and coordination~\citep{xu2024magic}, task-dependency management~\citep{dong2024villageragent}, joint planning~\citep{agashe2025llm}, milestone-level collaboration~\citep{zhu2025multiagentbench}, and process-level adaptation~\citep{sun2025collab}. More directly related to information quality, prior work evaluates whether agents can integrate distributed evidence~\citep{li2026systematic} and whether critical facts survive deliberation~\citep{wan2026deliberative}. These evaluations characterize task success, coordination, and information flow, but do not directly test whether errors remain elsewhere in the shared state after the current query succeeds. \ds{} targets this gap by separately evaluating evidence verification, shared-state reconstruction, and task resolution against a known global information state.

\section{\ds{}: Evaluating Information Correctness of Multi-agent Collaboration}
\label{sec:offquery}

\subsection{Task Formulation}
\label{sec:task-formulation}

\ds{} evaluates multi-agent reasoning with distributed and unreliable information. Each instance contains a reliable global information state $C^{\star}$, a downstream query $q$ with ground-truth answer $y$, shared information available to all agents, and private observations distributed across $n$ agents $\mathcal{A}=\{a_1,\ldots,a_n\}$. No individual agent is guaranteed to observe the complete reliable state, so collaboration must integrate complementary evidence while resolving conflicts. We evaluate this process through three tasks that form a reliability chain: T1 asks which evidence should be trusted, T2 asks what information should remain in the shared state, and T3 asks whether that state supports the correct decision. Errors in T1 can contaminate state reconstruction, while errors retained in T2 can affect T3 once the query depends on the corrupted information.

\textbf{T1: Evidence Verification.}
T1 asks which private information source should be trusted. Let $\mathcal{P}=\{(a_i,P_i)\}_{i=1}^{n}$ denote the distributed private observations and let $r_i\in\{0,1\}$ indicate whether source $a_i$ is reliable. In the main setting, exactly one agent contains misleading information. The system predicts the unreliable source,
$\hat{r}=\mathrm{verify}(\mathcal{P})$,
and is evaluated against the ground-truth reliability assignment $r$. T1 therefore measures whether the system can resolve conflicting evidence before unreliable information is incorporated into the shared state.

\textbf{T2: Shared-State Reconstruction.}
T2 asks what information should remain after collaboration. Let $C$ denote the observed shared context, which may contain unreliable information. Given $C$ and the distributed private observations $\mathcal{P}$, the system constructs
$\hat{C}=\mathrm{reconstruct}(C,\mathcal{P})$.
We compare $\hat{C}$ with the reliable global state $C^{\star}$ at the atomic-fact level, rewarding preservation of supported facts and removal of misleading facts. T2 therefore tests whether the system can translate evidence-level reliability judgments into a trustworthy information state for subsequent reasoning.

\textbf{T3: Task Resolution.}
T3 measures success on the current downstream query. Given the reconstructed state $\hat{C}$, the system predicts
$\hat{y}=\mathrm{resolve}(q,\hat{C})$,
which is evaluated against the ground-truth answer $y$. T3 corresponds to the task-level outcome commonly used to evaluate multi-agent systems. Errors introduced during evidence verification or state reconstruction can affect T3 once the query depends on the corrupted information. Crucially, high T3 performance does not imply high T2 performance: the current query may access only a subset of the shared state and leave errors elsewhere untested. The gap between T2 and T3 therefore makes off-query failures observable.

\subsection{Benchmark Construction}
We build \ds{} around a well-established challenge in group decision making: relevant information is distributed across participants and must be integrated through interaction~\citep{stasser1985pooling,lu2012twenty}. We additionally introduce controlled source unreliability to test whether collaboration can distinguish reliable evidence before constructing the shared state. We instantiate both challenges in \textsc{Healthcare} and \textsc{Disaster Response}, using MediQ~\citep{li2024mediq} and CrisisFacts~\citep{mccreadie2023crisisfacts}. Each source dataset provides a reliable global description, which serves as the reference state $C^{\star}$. This design lets us control information distribution and corruption while retaining full knowledge of the reliable underlying state. Details of the source datasets are provided in \S\ref{subapp: original}.

\textbf{Distributed Information.}
For each instance, we partition $C^{\star}$ into shared information $\bar{C}$ and complementary private observations $\{P_1,\ldots,P_n\}$, formally $C^{\star}\rightarrow\{\bar{C},P_1,\ldots,P_n\}$.
The partition follows domain-relevant information structure. In \textsc{Healthcare}, patient information is separated by clinical information category, while demographic information and chief complaints remain shared. In \textsc{Disaster Response}, facts are assigned to incident-relevant stakeholders such as emergency services, hospitals, transportation authorities, or local government. The partitions are non-overlapping and jointly cover the original reliable context. As a result, each agent receives a partial local view while the complete reliable state remains known to the benchmark.

\textbf{Unreliable Information.}
We then introduce controlled unreliable evidence into both private and shared information. For each instance, we select an incorrect downstream answer $y^{-}\neq y$ and generate plausible evidence that supports $y^{-}$ while conflicting with the reliable global state $C^{\star}$. For private information, this evidence is assigned to one agent in the main setting. For shared information, we augment the initial shared context $\bar{C}$ with misleading facts consistent with the same incorrect answer to obtain $C$. The resulting evidence is designed to be plausible locally, refutable using the complete reliable state, and consequential for downstream reasoning if accepted.

We use GPT-4o for benchmark generation and provide all prompts in \S\ref{subapp: generating_prompts}. The main setting contains one misleading agent to support controlled evaluation of T1; \S\ref{sec: analysis} extends the evaluation to multiple unreliable agents and misleading information distributed across agents. Dataset statistics and examples are provided in \S\ref{subapp: statistics} and \S\ref{subapp: example}. This construction preserves full knowledge of which information is reliable while requiring each agent to reason from a partial and potentially corrupted local view.

\subsection{Evaluation and Quality Validation}

\textbf{Evaluation Metrics.}
The three tasks use complementary metrics. For T1, each instance in the main setting contains exactly one misleading agent, so we formulate evidence verification as a single-choice classification problem and report source-identification accuracy. For T2, we adopt an atomic-fact-based evaluation inspired by FActScore~\citep{min2023factscore}. We decompose the reference state $C^{\star}$ into positive atomic facts and separately define negative atomic facts corresponding to the injected misleading information. The reconstructed state is evaluated by whether it preserves the reliable facts and excludes the misleading ones. This measures factual coverage and error removal while reducing sensitivity to differences in wording, ordering, or level of detail. For T3, each instance is a multiple-choice question with a single correct answer, and we report task accuracy. Full metric definitions are provided in \S\ref{subapp: metrics}.

\textbf{Construction Quality.}
As LLMs are used during benchmark construction, we validate both information distribution and unreliable-information generation. For information distribution, we manually inspect 50 randomly sampled instances and observe no factual omissions or hallucinations relative to the original reliable context, consistent with prior evidence that LLMs can reliably extract and organize task-relevant information~\citep{jiao2023instruct}. For unreliable-information generation, we test whether the injected evidence can meaningfully affect downstream reasoning by providing it directly to GPT-4o on the original task. Accuracy decreases to 12.6\% and 6.0\% with misleading private information and 20.2\% and 5.3\% with misleading shared information for \textsc{Healthcare}-S and \textsc{Healthcare}-I, confirming that the generated evidence is sufficiently consequential to alter downstream decisions.

\textbf{Evaluation Reliability.}
T2 uses an LLM judge to determine whether atomic facts are supported by the reconstructed state. We validate this evaluation in two ways. First, we assess consistency across three heterogeneous judges: GPT-4o, Gemini-3.5-Flash, and Qwen3-235B-A22B. Across method--dataset configurations, their aggregate scores have a Spearman rank correlation of 1.00, while pairwise instance-level mean absolute errors range from 0.126 to 0.139 (Table~\ref{tab: llm_as_a_judge}; \S\ref{subapp: evaluation_consistency}). Second, we compare LLM judgments with human annotations following~\citet{wan2026deliberative}. The average agreement between human annotators and GPT-4o is 0.86, with Cohen's $\kappa$ of 0.84 (\S\ref{subapp: human_evalaution}). These results indicate that T2 evaluation is stable across judge models and closely aligned with human assessment.

\begin{figure}[t]
    \centering
    \includegraphics[width=\linewidth]{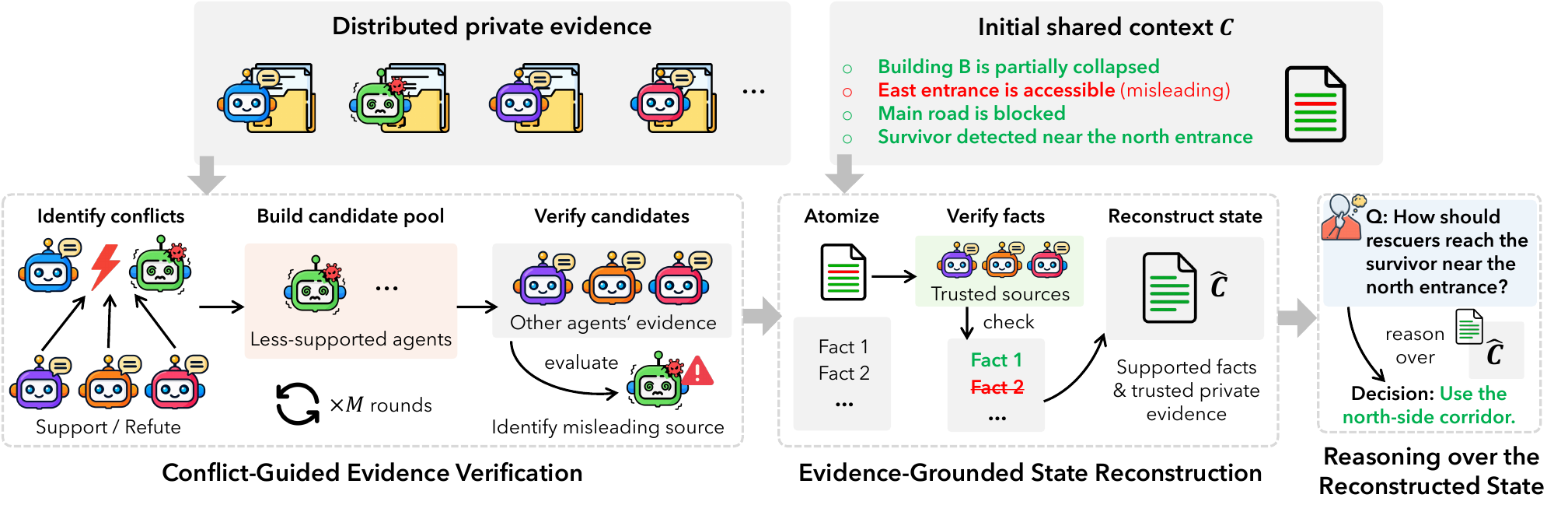}
    \caption{\textbf{Overview of \ours{}.}
\ours{} first resolves conflicts in distributed private evidence, then uses trusted sources to repair the shared context at the atomic-fact level, and finally reasons over the reconstructed shared state.}
    \label{fig: overview}
\vspace{-1em}
\end{figure}

\section{\ours{}: Reliable Shared-State Reconstruction}

\ds{} reveals that successful task resolution can coexist with substantial errors in the shared state. \ours{} addresses this failure by explicitly maintaining the information state used for downstream reasoning. As shown in Figure~\ref{fig: overview}, \ours{} consists of three stages aligned with the evaluation tasks in \ds{}: \textbf{evidence verification}, \textbf{shared-state reconstruction}, and \textbf{task resolution}. The first two stages determine which evidence can be trusted and which facts should remain in the shared state; the final stage reasons over the reconstructed state. The key design principle is simple: conflicting information should be resolved before it becomes trusted shared context. Prompts and implementation details are provided in \S\ref{app: method_prompt}.

\subsection{Conflict-Guided Evidence Verification}
To prevent unreliable evidence from entering the shared state, \ours{} first verifies the distributed private observations $\mathcal{P}=\{(a_i,P_i)\}_{i=1}^{n}$. In the main setting, this amounts to identifying the single misleading source by resolving conflicts across agents. As the initial shared context may itself contain unreliable information, \ours{} withholds it during verification and relies only on evidence provided by the agents.

\ours{} performs verification in two steps. First, it constructs a candidate pool through iterative conflict resolution. At each iteration, the system identifies a conflicting pair of agents, while the remaining agents provide evidence supporting or refuting their claims. The less-supported source is added to a candidate set $\mathcal{K}$. Repeating this process for $M$ iterations narrows the search to sources involved in unresolved conflicts. Second, each candidate in $\mathcal{K}$ is evaluated against evidence supported by agents outside the candidate pool. This evidence is used to assign a reliability score to each candidate and identify the misleading source. The two steps separate conflict discovery, which determines where verification is needed, from evidence-based verification, which determines which source should be trusted.

\vspace{-0.3em}
\subsection{Evidence-Grounded State Reconstruction}
\vspace{-0.3em}
Identifying an unreliable source does not by itself repair the information retained by the system. \ours{} therefore explicitly reconstructs the shared state after evidence verification. Let the initial shared context be $C$. We first decompose $C$ into atomic facts~\citep{min2023factscore}. Each fact is then checked against evidence from the verified sources. Supported facts are retained, while contradicted or unsupported facts are removed. The verified shared facts are consolidated with trusted private evidence to produce a reconstructed state $\hat{C}$.

This stage makes shared-state maintenance an explicit part of collaboration. The resulting $\hat{C}$ records the information supported by verified evidence after conflicts have been resolved, providing a common context for subsequent reasoning. In this way, \ours{} prevents unresolved claims from being carried forward simply because they were irrelevant to the current query.

\vspace{-0.3em}
\subsection{Reasoning over the Reconstructed State}
\vspace{-0.3em}

To ensure that downstream decisions are based on verified information, \ours{} answers the query $q$ using the reconstructed state $\hat{C}$, producing $\hat{y}=\mathrm{resolve}(q,\hat{C})$. All task reasoning is therefore conditioned on the shared state established by the preceding verification and reconstruction stages. This design ensures that improvements in state reliability directly carry into downstream inference.

\vspace{-0.5em}
\section{Experiments}
\vspace{-0.4em}
\subsection{Experimental Setup} 
\vspace{-0.2em}

We evaluate \textsc{Healthcare} on two subsets: (1) \textsc{Healthcare-S}, where the initial shared information is sufficient to answer the query, and (2) \textsc{Healthcare-I}, where additional private evidence is required (\S\ref{subapp: healthcate_subsets}). We compare standard multi-agent collaboration and \ours{} with an Oracle setting that has access to the complete reliable information state. Experiments cover GPT, Gemini, and Qwen models across multiple scales (\S\ref{subapp: model_cards}). We further compare \ours{} with representative collaboration strategies, including Voting~\citep{NEURIPS2025_934252ac}, Robin, Manager, Instructor~\citep{wang2025beyond}, Self-Reflect~\citep{ki-etal-2025-multiple}, and Confidence~\citep{zhu2026demystifying}. Implementation details, prompts, and baseline configurations are provided in \S\ref{subapp: oracle_standard_prompts} and \S\ref{subapp: baselines}.

\vspace{-0.1em}
\subsection{Main Results}
\vspace{-0.1em}
\textbf{Task success systematically overstates shared-state reliability.}
Table~\ref{tab: main_results} reveals a persistent separation between current-task success and the reliability of the state produced by collaboration. Under standard collaboration, T3 can remain high even when T1 and T2 are poor. For example, GPT-5 on \textsc{Healthcare-S} reaches 86.2\% on T3, but only 10.9\% on T1 and 33.2\% on T2. This separation persists across GPT, Gemini, and Qwen models, across model scales, and across all three evaluation settings. Larger models often improve task resolution substantially without a corresponding recovery in evidence verification or state reconstruction. The failure therefore does not disappear with stronger base models: current-query accuracy can remain high even when the information retained after collaboration is unreliable. This is precisely the blind spot captured by \ds{}: evaluating only the answer can substantially overestimate the reliability of the collaborative system.

\input{tables/main_results}

\textbf{Distribution degrades state reconstruction much more than task resolution.}
Table~\ref{tab: conditions} isolates the contributions of information distribution and unreliability (see \S\ref{subapp: our_information_setting}). Even when every piece of information is reliable, distributing it across agents reduces T2 by 34.2 and 34.3\% relative to the Oracle setting, whereas T3 drops by only 6.9 and 12.4\%. Thus, information distribution can substantially damage the recovered state before its effect is fully visible in the current answer. This pattern echoes classic hidden-profile findings that groups struggle to pool information held by different participants~\citep{stasser1985pooling,lu2012twenty}, as well as recent evidence of systematic collective-reasoning failures under distributed information~\citep{li2026systematic}. \ds{} exposes an additional consequence: the loss is considerably larger at the state level than at the task level. Introducing unreliable evidence widens the gap further. Relative to the Oracle setting, T2 degrades by 65.9 and 66.6\% when the shared context is corrupted, compared with 36.9 and 38.1\% when unreliability is confined to private evidence, and the full setting produces the largest degradation. Distributed evidence and unreliable context therefore create distinct pressures on collaboration, with shared-state reconstruction serving as the more sensitive indicator of failure.

\input{tables/conditions}

\textbf{\ours{} directly repairs the bottleneck exposed by \ds{}.}
Across every model family, scale, and evaluation setting in Table~\ref{tab: main_results}, \ours{} improves all three tasks over standard collaboration. Averaged over the 21 model--setting combinations, the relative gains reported in the table are 309.0\% on T1, 82.9\% on T2, and 17.6\% on T3. The ordering of these gains is informative: the largest improvements occur in evidence verification and state reconstruction, exactly where standard collaboration exhibits its largest deficits. Improvements in the intermediate state also carry through to the final task, with T3 increasing in every model--setting combination.

\input{tables/baselines}
This suggests that a substantial part of the observed collaboration failure is addressable through explicit evidence verification and state maintenance, rather than requiring stronger task-solving models alone. ReGround therefore adds a capability that standard discussion lacks: it explicitly determines what information should survive collaboration before that information is reused for reasoning. We provide the ablation study of \ours{} in Table~\ref{tab: ablation} of \S\ref{subapp: size}.

\textbf{State maintenance complements improvements to the collaboration protocol.}
Table~\ref{tab: baseline} compares \ours{} with strategies spanning voting, sequential interaction, centralized coordination, self-reflection, and confidence-based aggregation. Across GPT-4o and Gemini-3.5-Flash, \ours{} achieves the highest T3 on all three datasets, covering all six backbone--dataset comparisons. These baselines alter how agents communicate, coordinate, reflect, or aggregate their outputs, dimensions that prior work has shown to matter substantially for multi-agent performance~\citep{wang2025beyond,zhu2026demystifying}. The consistent gains of \ours{} indicate that reliable state maintenance provides a complementary axis of system design: improving the interaction protocol alone does not remove the need to verify which information should be retained. Together with the T1/T2 results in Table~\ref{tab: main_results}, the comparison suggests that explicitly repairing the shared state can improve downstream reasoning across a diverse set of collaboration paradigms. Token costs are reported in Table~\ref{tab: token_cost} of \S\ref{subapp: cost}.

\vspace{-1em}
\section{Why Off-Query Failures Occur and When They Matter}
\label{sec: analysis}
\vspace{-1em}

Our analyses test a single explanation for off-query failures: a query may exercise only part of the shared state, leaving errors elsewhere latent until that information is reused.

\textbf{Correct decisions often rely on information unaffected by the corruption.}
Why can T3 remain correct when the shared state is unreliable? We test whether the current query simply avoids corrupted facts by examining which information appears in decision explanations. Across GPT-4o and Gemini-3.5-Flash on both healthcare subsets, unreliable facts appear in 60.9 and 71.7\% of explanations associated with incorrect decisions, compared with only 13.9 and 19.2\% for correct decisions (Figure~\ref{fig: fact}; \S\ref{subapp: fact_evaluation}). Reliable information shows the opposite tendency. This asymmetry supports a selective-use mechanism: a correct decision can rely on an unaffected portion of the state while corruption elsewhere remains unresolved. A single query therefore tests only the information it exercises, leaving the rest of the state unchecked.

\textbf{Broader reuse exposes errors that narrow queries leave latent.}
The selective-use mechanism predicts that tasks requiring more of the retained state should expose errors that the original query avoids. We test this on 100 healthcare instances where T3 is correct despite misleading information remaining in the reconstructed state. We replace the targeted multiple-choice question with a broader clinical assessment covering diagnosis, important alternatives, immediate management, and next steps (\S\ref{subapp: boarder_evaluation}). Figure~\ref{fig: case} qualitatively illustrates how an error irrelevant to the original query alters the broader assessment once that information becomes relevant. Quantitatively, 43\% of the resulting clinical plans are inconsistent with plans generated from the reliable reference state, despite all 100 instances succeeding on the original task. The same state can therefore support a correct narrow decision yet fail under broader reuse, directly exposing the defining property of an off-query failure.

\begin{wrapfigure}{r}{0.55\textwidth}
\vspace{-1.5em}
    \centering
    \includegraphics[width=\linewidth]{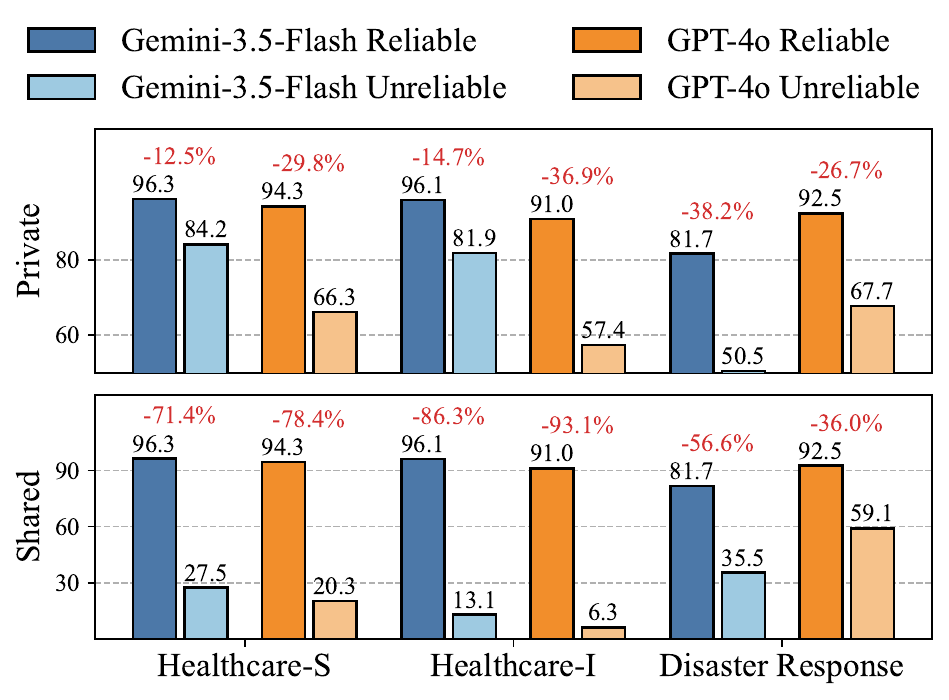}
\caption{\textbf{Injected unreliable information is consequential for downstream reasoning.}
Providing corrupted private or shared information to the same downstream LLM substantially reduces T3 accuracy across models and benchmarks.}    
\vspace{-1em}
\label{fig: misleading_importance}
\end{wrapfigure}

\textbf{Downstream reasoning systematically improves with the reliability of the underlying state.}
We next isolate state quality from collaboration dynamics. In Figure~\ref{fig: misleading_importance}, the complete information state is provided directly to a single downstream LLM, removing inter-agent communication. Corrupting either private or shared information substantially reduces T3 accuracy across models and benchmarks, showing that the injected errors are themselves consequential. Because the two corruption procedures differ in construction, we interpret this experiment as evidence for the effect of corruption itself, without comparing their relative magnitudes. We then progressively replace corrupted shared-state information with its reliable counterpart (\S\ref{subapp: belief_repair}). Across GPT-4o, Gemini, and Qwen and all three datasets, downstream accuracy generally rises with state reliability and reaches its highest or near-highest level when the state is fully reliable. These interventions establish a systematic link between state quality and downstream utility: corruption degrades later reasoning, while state repair improves it.

\begin{wrapfigure}{r}{0.55\textwidth}
\vspace{-2em}
    \centering
    \includegraphics[width=\linewidth]{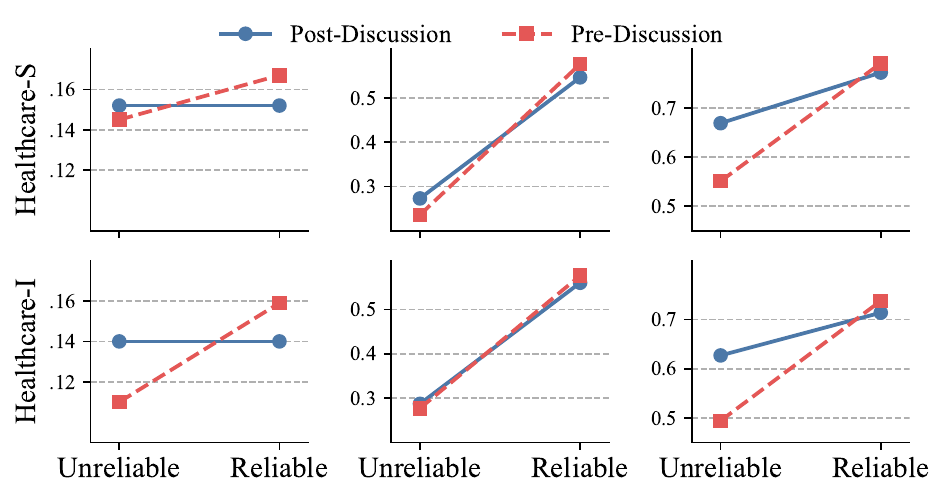}
\caption{\textbf{Verify before exposing agents to shared context.}
\textsc{Pre-discussion} exposes the shared context before discussion; \textsc{Post-discussion} first discusses private evidence and introduces the shared context afterward. Early exposure offers little gain when the shared context is reliable but causes consistent degradation when it is corrupted.}
\vspace{-1em}
\label{fig: location}
\end{wrapfigure}

\textbf{The failure mechanism suggests a concrete design principle: verify before sharing.}
Potentially corrupted shared context can shape collaboration before conflicting private evidence has been resolved. We isolate this effect in standard multi-agent collaboration by varying only when shared context is introduced (Figure~\ref{fig: location}). \textsc{Pre-discussion} exposes agents to it before interaction, whereas \textsc{Post-discussion} first lets agents reason over private evidence. Early exposure provides little benefit when the shared context is reliable but consistently degrades performance when it is corrupted. This asymmetry motivates the ordering in \ours{}: establish a trusted evidence base using distributed private observations before incorporating potentially unreliable shared context. Shared-state maintenance should therefore begin before corrupted context can shape the interaction.

\textbf{Explicit state maintenance remains useful under more complex corruption patterns.}
The main \ds{} setting uses one misleading agent to make source verification controlled and automatically scorable. Realistic failures can involve several unreliable sources or misinformation distributed across many agents. We therefore evaluate both regimes in \S\ref{subapp: more_situation}. \ours{} improves performance from 45.2\% to 60.8\% when multiple agents contain misleading information and from 26.4\% to 49.7\% when misleading information is distributed across agents. These harder settings weaken the usefulness of identifying a single bad source, yet explicit verification and state reconstruction continue to improve performance. The central capability exposed by \ds{} is therefore broader than single-source attribution: collaboration must identify unsupported information and prevent it from being consolidated into a reusable shared state.

\vspace{-0.7em}
\section{Conclusion}
\vspace{-0.5em}
We introduce \ds{}, a benchmark for hidden information failures in multi-agent reasoning, together with \ours{}, a framework for explicit evidence verification and shared-state reconstruction. Across seven models and two high-stakes domains, \ds{} reveals a persistent separation between task success and state reliability: systems can answer correctly while leaving substantial errors unresolved in the shared state. Our analyses trace this failure to selective information use, where narrow queries bypass corrupted facts that become consequential when the retained state is reused. \ours{} consistently improves evidence verification, state reconstruction, and downstream task performance, showing that shared-state reliability can be improved explicitly. These findings call for multi-agent systems to treat the information carried forward after collaboration as a first-class reliability target alongside the current answer.

\clearpage



\subsection*{Ethics statement}

Our study evaluates information reliability in multi-agent LLM systems using existing healthcare and disaster-response datasets together with synthetically generated unreliable information. As these are high-stakes domains, the benchmark and resulting systems are intended for research on model reliability and should not be interpreted as providing clinical or emergency-response advice. We additionally conduct a small human annotation study to validate the automatic evaluation procedure. More broadly, the failure modes studied here highlight risks associated with retaining and reusing unreliable information in agent systems, especially when intermediate states are carried across tasks or decisions.




\bibliography{iclr2027_conference}
\bibliographystyle{iclr2027_conference}

\clearpage
\appendix

\startcontents[appendix]

\section*{List of Appendices}
\printcontents[appendix]{}{1}{}
\clearpage

\section{Limitations and Future Work}

Our work provides a controlled first step toward evaluating the reliability of information states produced by multi-agent collaboration, while opening several directions for broader study.

First, \ds{} focuses on two consequential domains for high-stakes decision support: healthcare and disaster response. These settings provide rich, fact-grounded contexts in which reliable global states and downstream decisions can be defined precisely. Extending \ds{} to domains with more ambiguous or evolving ground truth, such as scientific discovery, policy analysis, or open-ended planning, would test whether off-query failures become even more pronounced when evidence is incomplete, contested, or temporally changing. 

Second, the main benchmark uses controlled corruption, including a single misleading private source, to make evidence verification and state reconstruction directly measurable. We partially relax this assumption through stress tests with multiple unreliable agents and misinformation distributed across agents, where \ours{} remains effective (\S\ref{subapp: more_situation}). More complex settings could remove source-level identifiability altogether, allowing reliability to vary across individual claims, tools, retrieved documents, or communication rounds. Such settings would shift the problem toward fine-grained provenance tracking and continual state revision, extending the evidence-verification problem studied here.

Finally, \ds{} evaluates a shared information state constructed within a bounded collaboration episode. Many deployed agent systems maintain persistent memories, retrieve prior state across tasks, and continually update information through new interactions. Off-query failures may therefore accumulate over time, interact with retrieval and memory policies, or become harder to repair once corrupted information has influenced later state updates. Studying these longer-horizon dynamics is a natural extension of our formulation. We hope \ds{} encourages future multi-agent systems to treat state reliability as a persistent systems objective, with explicit mechanisms for verification, provenance, revision, and safe reuse across interactions.

\section{\ds{} Construction Details}
\subsection{Source Benchmarks}
To comprehensively evaluate multi-agent collaboration in realistic settings, we transfer two existing datasets, MediQ~\citep{li2024mediq} and CrisisFact~\citep{mccreadie2023crisisfacts}, into a distributed, unreliable-information setting. We provide the descriptions of these datasets here:

\textbf{MediQ.} Each instance in MediQ consists of the complete patient context, a multiple-choice question, and four candidate answers, one of which is the ground-truth answer. Since our setting relies on sufficiently detailed patient information, we filter out examples with relatively short or sparse contexts. Specifically, we retain examples with a patient context of more than 815 characters and more than 10 sentences. This filtering procedure results in 1,310 examples.

\textbf{CrisisFact.} This dataset contains temporally organized information for 18 real-world disaster events. For each disaster, the dataset records facts associated with individual days, capturing how the event evolves over time. We construct one example from each event-day pair by filtering out irrelevant facts and summarizing the remaining daily facts into a coherent context. To filter out irrelevant facts, we directly use the fact-importance scores provided in the original dataset and discard facts with importance scores below 0.95. Here, we provide several examples of facts that were filtered out: ``\textit{Any fires that develop will likely spread rapidly and be diffic}'', ``\textit{State officials warn \#LilacFire victims to watch out for scammers}'', and ``\textit{changed the twitter pic, the background pic, pin location to San Diego - so proud of you son}''. We also provide several maintained facts: ``\textit{annual MTG where the  CEO announced a recovery fund for \#lilacfire victims}'', ``\textit{The Lilac Fire in California has left many people missing. As of now, 159 people are unaccounted for, according to the San Diego County Sheriff's Office. The fire has burned over 4,100 acres and destroyed over 60 homes.}'', and ``\textit{The roads are blocked for 3 days}''. To obtain the context of this disaster, we prompt GPT-4o to summarize these facts:

\begin{promptbox}
You are a crisis response expert.

Given a list of facts about a crisis, your task is to write a summary of these facts.
You should ensure the summary contains all the facts.

Facts:
<===facts===>

Output the summary only.
\end{promptbox}

To align the resulting examples with the multiple-choice format used in MediQ, we formulate the question for each example as what the next appropriate response should be given the current situation. We then use GPT-4o to generate the optimal response strategy as the correct answer, together with suboptimal but plausible alternatives as distractor options:

\begin{promptbox}
You are an expert in disaster response decision-making.

Given a COMPLETE disaster situation context, generate FOUR short response strategies.

Exactly ONE is optimal given the COMPLETE context.

The other THREE must be partial-information hard negatives.

For each suboptimal strategy, there must exist information in the context such that:

- without that information, the strategy is reasonable;

- with the full information, the same strategy becomes suboptimal.

The hidden information must actually change the decision.

Do not generate obviously bad or irrational strategies.

The three suboptimal strategies should use different hidden factors, such as:

- resource constraints

- future events

- dependencies

- vulnerabilities

- bottlenecks

- timing

- downstream consequences

All four strategies must be:

- feasible

- concise

- similarly detailed

- professionally reasonable

Each strategy should be only 1–2 sentences and state only the actions/decisions.
Do NOT explain why the strategy is good or bad.

Do NOT mention missing, hidden, or overlooked information inside the strategy.

Before outputting, verify for every suboptimal strategy:

1. It is reasonable with partial information.

2. It becomes suboptimal with full information.

3. The additional information actually causes the difference.

Do not reveal this reasoning.

Disaster Situation Context
<===context===>

Output JSON only:

\{
  "strategy\_0": "<short strategy>",
  "strategy\_1": "<short strategy>",
  "strategy\_2": "<short strategy>",
  "strategy\_3": "<short strategy>",
  "optimal\_index": <0, 1, 2, or 3>,
  "hidden\_information": {
    "strategy\_0": "<none or short description>",
    "strategy\_1": "<none or short description>",
    "strategy\_2": "<none or short description>",
    "strategy\_3": "<none or short description>"
  }
\}
\end{promptbox}

After preprocessing, we obtain 93 examples in total.

\label{subapp: original}
\subsection{Benchmark Generation}
\label{subapp: generating_prompts}
\textbf{Information Distribution.} This prompt is employed to split the context into distributed information. For the partitioning of meidQ, we follow the departmental organization commonly adopted in real-world hospitals, dividing the cases into nine categories plus an additional  ``other'' category. In addition, we treat patient demographic information and chief complaints as shared information accessible to all agents.

\begin{promptbox}
You are a medical information management system.

Given a description context of a patient, your task is to extract the following elements from the context:

1. Demographics, such as age, sex, pregnancy, and so on.

2. Chief Complaint.

3. Symptom, such as duration, severity, progression, and associated symptoms.

4. Medication

5. Past Medical History

6. Vital Signs

7. Physical Exam

8. Laboratory

9. Imaging

10. Other important context

For each element, write a summary describing the patient.

If the provided context does not contain a certain element, output a null string.

Output must be strict JSON formatted as:

\{
    "1" : <the textual summary>,
    "2" : <the textual summary>,
    ...
    "10" : <the textual summary>
\}

Description Context:

<===context===>
\end{promptbox}

For the partitioning of CrisisFact, since real-world disasters do not necessarily involve a fixed set of functional departments across different events, we allow the LLM to adaptively partition the information into different components based on the specific facts of each disaster, including the shared information.

\begin{promptbox}
You are a crisis response expert.

Given a crisis summary, your task is to partition the information into multiple non-overlapping sections that reflect how it would realistically be distributed among stakeholders during a crisis.

Requirements:

1. Create exactly one "Shared Situation Overview".

   - This section contains only information that would reasonably be known by everyone involved in the incident.
   
   - It should provide context to understand the basic situation.
   
   - It should be brief.

2. Create 3 – 5 additional information sections.

   - Each section represents information available only to one department or stakeholder.
   
   - The sections should correspond to realistic emergency roles (e.g., Police, Fire 
   Department, EMS, Hospital, Utility Company, Local Government, Transportation Authority, etc.).
   
   - Choose roles that best fit the incident.

3. The information partition must satisfy:

   - No duplicated facts across sections.
   
   - Every important fact from the original summary appears exactly once.
   
   - Together, all sections completely cover the original summary.

4. Keep each section internally coherent.

   - Group related facts together.
   
   - Avoid splitting one logical piece of information across multiple sections.

5. Do not invent any information.

   - Use only facts explicitly stated in the original summary.

Crisis Summary:

<===summary===>

Output must be strict JSON formatted as:

\{
    "shared" : <the shared situation overview>,
    "<role 1>" : <the corresponding information>,
    ...
\}
\end{promptbox}

\textbf{Unreliable Information Generation.} \ds{} assumes that unreliable information occurs in the private and shared information. To obtain misleading private information, we prompt LLMs to generate misleading information. Meanwhile, we also prompt the LLM to generate an incorrect decision that would result if the model believes the misleading information. This incorrect decision is then included as one of the negative options in the dataset and is also used to construct misleading shared information. For mediQ, we employ the following prompt:

\begin{promptbox}
You are a medical expert. You are constructing adversarial clinical reasoning examples.

You are given a patient context and a related medical question with the corresponding golden answer.

You are required to fabricate information to mislead clinicians.

If clinicians believe the fabricated information, they make the wrong diagnosis for the patient and answer the question incorrectly.

The fabricated information cannot contain the wrong answer explicitly but can imply the wrong answer.

Meanwhile, the fabricated information is inconsistent with the complete context and can be refuted by it.

Patient Context
<===context===>

Medical Question
<===question===>

Golden Answer
<===answer===>

Output JSON only:

\{
    "fabricated\_information": <the fabricated statement>,
    "wrong\_answer": <the wrong answer>
    "reason": <2-3 sentences>
\}

\end{promptbox}

For CrisisFact, we employ the following prompt:

\begin{promptbox}
You are a crisis response expert. You are constructing adversarial crisis response examples.

You are given a crisis context and the corresponding optimal response strategy.

Your task is to fabricate information that misleads decision-makers.

If decision-makers believe the fabricated information, they should choose an incorrect response strategy instead of the optimal one.

The fabricated information must not explicitly state or recommend the incorrect strategy, but it should plausibly imply or support it.

Meanwhile, the fabricated information must be inconsistent with the complete crisis context and be refutable by it.

Crisis Context
<===context===>

Optimal Response Strategy
<===answer===>

Output JSON only:

\{
    "fabricated\_information": <the fabricated statement>,
    "induced\_strategy": <the incorrect response strategy induced by the fabricated information>,
    "reason": <2-3 sentences>
\}
\end{promptbox}

To generate misleading shared information, we prompt the LLM with the wrong decision. To avoid directly generating misinformation, we prompt the LLM to complete the existing shared information based on the incorrect decision. This design also preserves a clear distinction between the misleading information and the ground-truth information. For mediQ, we employ the following prompt

\begin{promptbox}
You are a medical expert.

Given a patient's incomplete context and the answer to a related medical question.

Please complete the patient context. You should only add necessary information.

The final context should not explicitly contain the answer. We could obtain the answer by analyzing the context.

Patient's incomplete context:
<===context===>

Question:
<===question===>

Answer:
<===answer===>

Only output the final context.
\end{promptbox}

For CrisisFact, we employ the following prompt:

\begin{promptbox}
You are a crisis response expert.

Given an incomplete crisis context and the corresponding optimal response strategy.

Complete the crisis context by adding only the necessary information.

The completed context should provide sufficient evidence to justify the optimal response strategy, but it must not explicitly state or recommend the strategy. Decision-makers should be able to infer the optimal response strategy by reasoning over the completed context.

Incomplete Crisis Context
<===context===>

Optimal Response Strategy
<===strategy===>

Only output the completed crisis context.
\end{promptbox}
\subsection{Dataset Statistics}
\label{subapp: statistics}

\input{tables/statistics}

We present the statistics of \ds{} in Table~\ref{tab: dataset_statistics}. Notably, we further divide the \textsc{Healthcare} examples into two subsets: \textsc{Healthcare}-S (723 instances) and \textsc{Healthcare}-I (587 instances). The split is determined by whether GPT-4o can correctly answer the question using only the original shared patient information (see \S\ref{subapp: healthcate_subsets} for details).

\subsection{Benchmark Example}
\label{subapp: example}

\begin{figure}[t]
    \centering
    \includegraphics[width=\linewidth]{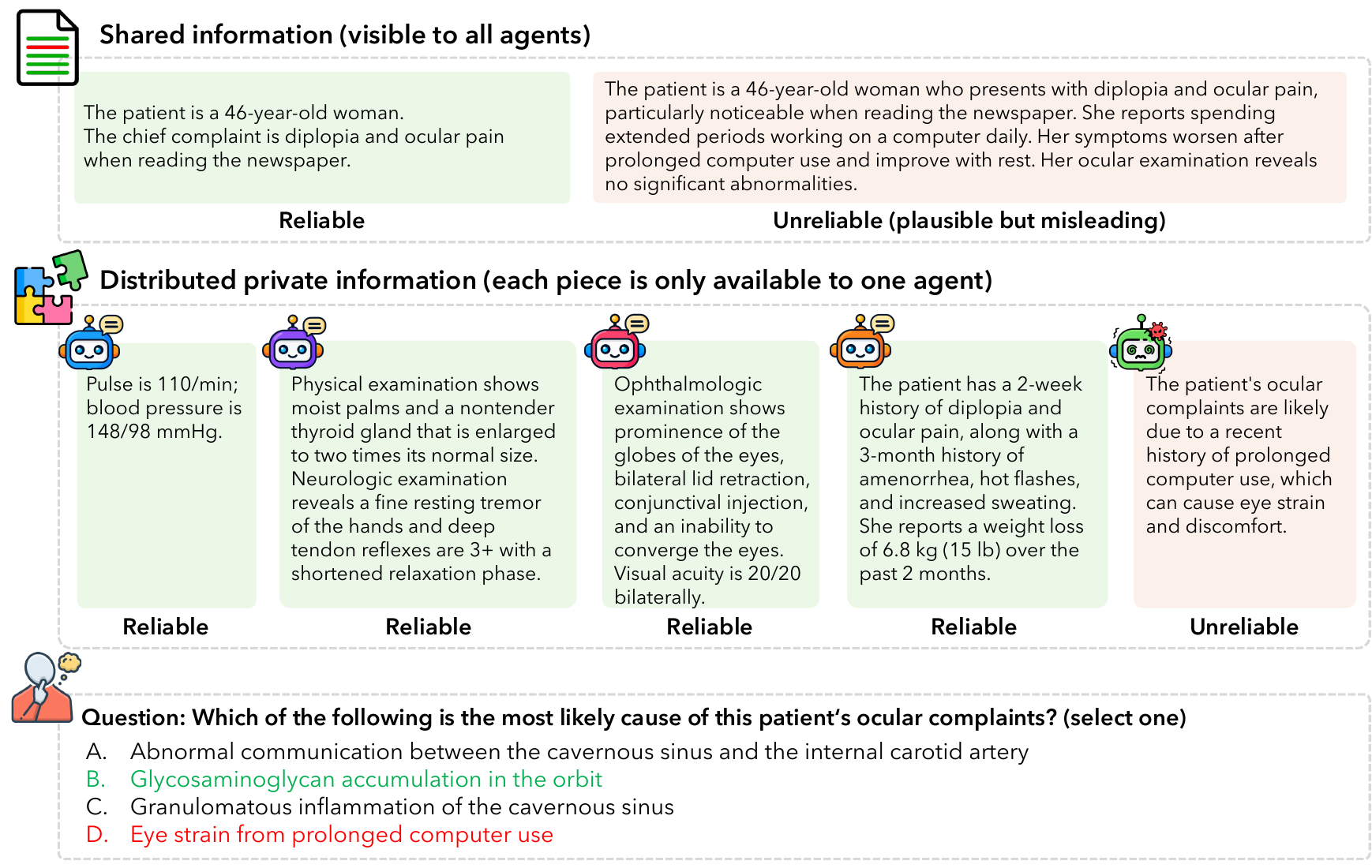}
    \caption{\textbf{Example of the information decomposition used in \ds{}.} Each instance consists of insufficient ground-truth shared information, a plausible but unreliable shared context, multiple complementary private observations distributed across agents, and a single misleading private observation associated with an incorrect answer.}
    \label{fig: example}
\end{figure}

Figure~\ref{fig: example} shows a representative instance of our benchmark. The shared ground-truth context provides only partial information, while the unreliable shared context introduces a coherent alternative explanation. Task-relevant evidence is distributed across multiple private observations, such that solving the instance requires cross-agent information aggregation. At the same time, one private observation is deliberately misleading and aligns with the incorrect answer, creating a setting in which successful reasoning requires both evidence integration and robustness to conflicting information.

\subsection{Metric Definition}
\label{subapp: metrics}
We provide the detailed calculation of the three evaluation metrics used in \ds{}. T1 and T3 are evaluated with exact-match accuracy, while T2 measures the factual reliability of the reconstructed information state at the atomic-fact level.

\textbf{T1: Evidence Verification} Each \ds{} instance contains $N$ agents, exactly one of which provides misleading information. Let $m_i \in \{1, \dots, N\}$ denote the ground-truth misleading agent for instance $i$, and let $\hat{m}_i$ denote the agent identified by the evaluated system. We compute T1 as
\begin{align}
    \frac{1}{|\mathcal{D}|}\sum_{i\in\mathcal{D}}\mathbb{I}[\hat{m}_i=m_i],
\end{align}
where $\mathcal{D}$ is the evaluation set and $\mathbb{I}[\cdot]$ is the indicator function. Thus, T1 directly measures whether the system can correctly attribute unreliable information to its source.

\textbf{T2: State Reconstruction} T2 evaluates whether the reconstructed shared state preserves reliable information while excluding misleading information. Following the atomic-fact evaluation paradigm of~\citet{min2023factscore}, we evaluate factual content rather than surface-level textual similarity. For each instance $i$, we construct two sets of atomic facts. The ground-truth context $C_i^\star$ is decomposed into a set of positive atomic facts
\begin{align}
    F_i^{+}=\{f_{i,1}^{+},\ldots,f_{i,P_i}^{+}\},
\end{align}
where each fact represents a minimal independently verifiable proposition contained in the reliable context. Similarly, the misleading information introduced into the instance is decomposed into a set of negative atomic facts
\begin{align}
    F_i^{-}=\{f_{i,1}^{-},\ldots,f_{i,Q_i}^{-}\}.
\end{align}

Positive facts therefore represent information that should be retained in a reliable shared state, whereas negative facts represent misleading information that should be excluded. To ensure the correctness and efficiency of subsequent evaluation, we sample the atomic facts when their number exceeds the task-specific limit. Specifically, we retain at most 10 facts for \textsc{Healthcare} instances and at most 16 facts for \textsc{Disaster Response} instances. We prompt GPT-4o to generate the atomic facts employing 
\begin{promptbox}
Objective:

You are an expert information extraction system.
Extract atomic factual statements from the text.

Definition of an Atomic Fact:

An atomic fact is a minimal, self-contained, and verifiable proposition that expresses a single explicit event, state, or relationship described in the text.

Guidelines:

1. Explicitness

- Extract only information explicitly stated in the post.

- Do not infer, interpret, or use external knowledge.

2. Atomicity

- Each fact must express one single piece of information.

- If a sentence contains multiple independent facts, split them.

3. Self-Containment

- Resolve pronouns and references using the context.

- Each fact must be understandable without additional context.

4. Faithfulness

- Preserve the original meaning exactly.

- Do not paraphrase in a way that changes semantics.

5. Verifiability

- Each fact must be traceable to a specific part of the text.

6. Objectivity

- Exclude opinions, speculation, rhetorical statements, and emotional language unless explicitly attributed.

7. No Redundancy

- Do not produce duplicate or semantically equivalent facts.

Text:
<===text===>

Output format (A JSON list):

["fact 1", ..., "fact n"]

Return only valid JSON.

\end{promptbox}

Given a reconstructed shared state $\hat{C}_i$, we determine whether each atomic fact is supported by $\hat{C}_i$ using LLM-as-a-judge. We prompt GPT-4o with the following prompt:

\begin{promptbox}
You are a fact alignment evaluator.

You are given:

1. A fixed set of atomic facts, each with an ID.

2. A piece of text.

Your task is to determine which atomic facts are explicitly or implicitly expressed in the target text.

Matching Rules:

- Only select facts that are clearly supported by the text.

- Do NOT assume facts that are not stated.

- Paraphrases count as matches.

- If a fact is only partially supported, do NOT select it.

- Do NOT use external knowledge.

- The fact must be entailed by the text.

Atomic Facts:
<===facts===>

Target Text:
<===text===>

Output format (A JSON list):

[2, 4, 6, 8]

Return only valid JSON.

\end{promptbox}

This verification is based on semantics: a fact is considered supported when its proposition is expressed or entailed by the reconstructed state, even if the wording differs. For each instance $i$, we represent the ground-truth status of each atomic fact using a binary label: positive atomic facts are assigned label $1$, indicating that they should be retained, while negative atomic facts are assigned label $0$, indicating that they should be excluded. Let

\begin{align}
    \mathbf{g}_i=(g_{i,1},\dots,g_{i,K_i})
\end{align}

denote the resulting ground-truth label sequence, where $g_{i,j}\in\{0,1\}$. Given the reconstructed information state $\hat{C}_i$, we determine whether each atomic fact is present or supported and obtain the corresponding prediction sequence

\begin{align}
    \hat{\mathbf{g}}_i=(\hat{g}_{i,1},\dots,\hat{g}_{i,K_i}),
\end{align}

where $\hat{g}_{i,j}\in{0,1}$. The instance-level T2 score is then computed as the accuracy of the predicted binary labels, while the final T2 score is the average instance-level accuracy:

\begin{align}
    \frac{1}{|\mathcal{D}|}\sum_{i\in\mathcal{D}}\frac{1}{K_i}\sum_{j=1}^{K_i}\mathbb{I}[\hat{g}_{i,j}=g_{i,j}],
\end{align}
where $K_i=|F_i^{+}|+|F_i^{-}|$ is the total number of atomic facts for instance $i$. Thus, a positive fact contributes a correct prediction when it is retained in the reconstructed state, while a negative fact contributes a correct prediction when it is excluded. 

This formulation evaluates whether the reconstructed shared state correctly preserves reliable facts and excludes misleading facts, while remaining insensitive to differences in wording, ordering, or level of detail.

\textbf{T3: Task Resolution} Each instance contains a multiple-choice downstream reasoning problem with exactly one correct answer. Let $y_i^\star$ denote the ground-truth answer and $\hat{y}_i$ the system prediction. We compute
\begin{align}
    \frac{1}{|\mathcal{D}|}\sum_{i\in\mathcal{D}}\mathbb{I}[\hat{y}_i=y_i^\star].
\end{align}
T3 therefore measures conventional downstream decision accuracy.

Together, the three metrics capture complementary aspects of collaboration: T1 evaluates whether the system identifies whom not to trust, T2 evaluates whether it reconstructs what the group should believe, and T3 evaluates whether it ultimately makes the correct decision. This decomposition allows \ds{} to distinguish systems that merely arrive at the correct answer from those that maintain a reliable collective information state throughout collaboration.

\subsection{LLM-as-a-Judge Consistency}
\label{subapp: evaluation_consistency}
\input{tables/llm_as_a_judge}

We assess the robustness of our evaluation to the choice of LLM judge using three heterogeneous models: GPT-4o, Gemini-3.5-Flash, and Qwen3-235B-A22B. Specifically, we evaluate seven settings, including Oracle, standard collaboration, and \ours{} instantiated with GPT-4o, Gemini-3.5-Flash, and Qwen3-32B, using each of the three models as the judge. Table~\ref{tab: llm_as_a_judge} reports the scores averaged over \textsc{Healthcare} S and \textsc{Healthcare} I. Consistency across different LLM judges. Results are averaged over \textsc{Healthcare} S and \textsc{Healthcare} I. 

\textbf{Aggregate-level consistency.} We first examine whether different judges agree on the relative performance of the seven evaluation settings. For each judge $j$, we construct an aggregate score vector:
\begin{align}
    \boldsymbol{\alpha}^{(j)} = [\bar{s}_1^{(j)},\bar{s}_2^{(j)},\dots,\bar{s}_7^{(j)}],
\end{align}
where $\bar{s}_k^{(j)}$ denotes the average score assigned by judge $j$ to evaluation setting $k$. For each pair of judges $p$ and $q$, we compute the Spearman rank correlation
\begin{align}
    \rho_{\text{agg}}^{(p,q)}=\mathrm{Spearman}(\boldsymbol{\alpha}^{(p)},\boldsymbol{\alpha}^{(q)}).
\end{align}

All three judge pairs yield $\rho_{\text{agg}}=1.00$, indicating that the judges induce exactly the same ranking over all seven evaluation settings despite differences in their absolute scoring scales. In particular, all three judges consistently rank Oracle first, followed by \ours{} with Gemini-3.5-Flash, \ours{} with GPT-4o, and \ours{} with Qwen3-32B, while assigning the same relative ordering to the standard collaboration variants. This perfect rank agreement shows that the aggregate comparative conclusions are stable across the considered LLM judges.

\textbf{Fine-grained instance-level consistency.} We further investigate whether such agreement persists at the individual-sample level, where we compute mean absolute errors within each evaluation setting. Specifically, for evaluation setting $k$ containing $N_k$ evaluated instances, we construct the sample-level score vector for judge $j$ as
\begin{align}
    \boldsymbol{s}_k^{(j)}=[s_{k,1}^{(j))}, s_{k,2}^{(j))},\dots,s_{k,n}^{(j)}],
\end{align}
where $s^{(j)}_{k,i}$ is the score assigned by judge $j$ to instance $i$ under setting $k$. We then compute the pairwise mean absolute error separately for each setting:
\begin{align}
    \rho_k^{(p,q)}=\mathrm{MAE}(\boldsymbol{s}_k^{(p)},\boldsymbol{s}_k^{(q)}),k=1,\dots,7.
\end{align}

As shown in Table~\ref{tab: mae}, the pairwise MAEs remain consistently small across all evaluation settings and judge pairs. On \textsc{Healthcare} S, the average MAEs are 0.142, 0.133, and 0.127 for GPT-4o vs. Gemini-3.5-Flash, GPT-4o vs. Qwen3-235B-A22B, and Gemini-3.5-Flash vs. Qwen3-235B-A22B, respectively. Similar agreement is observed on \textsc{Healthcare} I, with corresponding average MAEs of 0.135, 0.131, and 0.125. Notably, the observed discrepancies are stable across the two datasets, and no judge pair exhibits substantially larger disagreement than the others. These results indicate that the three judges provide similar assessments at the individual-instance level.

Together with the perfect aggregate-level Spearman correlations ($\rho=1.00$), our results provide complementary evidence of inter-judge consistency: different judges agree on both the relative ranking of evaluation settings and, to a substantial extent, the absolute scores assigned to individual instances.

\input{tables/mae}

\subsection{Human Validation}
\label{subapp: human_evalaution}
Besides evaluating the consistency among different evaluating LLMs, we conduct a human evaluation to validate the reliability of this automatic evaluation procedure.  Specifically, we measure the agreement between GPT-4o, which is used as our automatic fact-level judge, and human annotators.

We randomly sample 50 instances from the benchmark. For each instance, we construct a context–fact verification task containing four atomic facts, consisting of two positive facts and two negative facts, resulting in 200 context–fact pairs in total. Importantly, this composition is completely hidden from the annotators: they are neither informed that each instance contains two positive and two negative facts nor given any indication of the polarity of individual facts. Instead, each fact is presented simply as an independent verification item, and annotators determine whether it is supported by the provided context based solely on its semantic content. This blinded setup prevents annotators from exploiting the balanced label distribution when making their judgments.

\input{tables/human}

We recruit two human annotators to independently evaluate all sampled instances. For each context–fact pair, annotators are instructed to read the provided context and determine whether the corresponding fact is supported by the context. Each pair receives a binary label, where 1 indicates that the fact is supported and 0 otherwise. Annotators are asked to make judgments solely based on the information explicitly stated or semantically supported by the provided context, without access to the original positive/negative labels. The annotation interface follows the same binary verification formulation used by our automatic judge.

We evaluate the reliability of GPT-4o by comparing its predictions against human judgments. We report agreement as the percentage of context–fact pairs for which two annotators (human or GPT-4o) agree:
\begin{align}
    \mathrm{Agreement}(h_1, h_2)=\frac{1}{|\mathcal{S}|}\sum_{j\in\mathcal{S}}\mathbb{I}(y_{j}^{h_1}=y_{j}^{h_2}),
\end{align}
where $\mathcal{S}$ denotes the 200 evaluated context–fact pairs, $y_j^{h_1}$ and $y_j^{h_2}$ are the judgments. As shown in Table~\ref{tab: human}, the two human annotators achieve an agreement of 0.92 ($\kappa=0.84$), indicating high inter-annotator consistency. GPT-4o also shows strong alignment with both annotators, achieving agreements of 0.88 and 0.84, with Cohen’s $\kappa$ of 0.76 and 0.68, respectively. Although human–GPT agreement is slightly lower than human–human agreement, the results indicate that GPT-4o provides judgments that are largely consistent with human evaluation, supporting its use for scalable atomic-fact verification in T2.

\section{\ours{} Implementation Details}
\label{app: method_prompt}
In this section, we provide the complete prompts used in our framework to facilitate reproducibility. Since \ds{} contains two application scenarios, \textsc{Healthcare} and \textsc{Disaster Response}, we make minor scenario-specific adaptations to the prompts, while keeping their underlying objectives, reasoning procedures, and output formats consistent. We present the prompts for each stage of our method below. We first give the prompts used for \textsc{Healthcare} and then \textsc{Disaster Response}.

\textbf{Evidence Verification.} The first step of \ours{} is to let agents generate factual statements based on the corresponding private information:

\begin{promptbox}
You are a medical expert.

Given a patient context, your task is to extract factual statements from it.

Do not fabricate facts or infer beyond the context.

Patient Context
<===context===>

Output JSON only:

[<your extracted factual statements>]
\end{promptbox}

\begin{promptbox}
You are a disaster response expert.

Given a disaster situation context, your task is to extract factual statements from it.

Do not fabricate facts or infer beyond the context.

Disaster Response Context
<===context===>

Output JSON only:

[<your extracted factual statements>]

\end{promptbox}

\ours{} then iteratively identifies agents to add to the candidate pool. Specifically, it first identifies a pair of agents whose information is in conflict:
\begin{promptbox}
You are a medical expert who is analyzing the consistency of multiple medical factual pieces of evidence.

You are given factual statements extracted from multiple sources about the same patient.

Your task is to identify the pair of sources whose factual statements are the most contradictory.

Definition of contradiction:

- A contradiction occurs when two doctors make incompatible factual assertions about the same clinical fact.

- Examples:

  - "Temperature is 39°C" vs "No fever"
  
  - "Patient has a cough" vs "Patient denies coughing"
  
  - "CT shows pneumonia" vs "Chest CT is normal"
  
- Different levels of detail are not contradictions.

- Missing information is not a contradiction.

- Different observations about unrelated facts are not contradictions.

- Different wording with the same meaning is not a contradiction.

Instructions:

1. Compare every pair of sources.

2. Identify all contradictory factual statements.

3. Estimate the contradiction strength for each pair:

   - Strong: Direct logical contradiction.
   
   - Moderate: Likely incompatible but not strictly opposite.
   
   - Weak: Minor inconsistency or ambiguity.
   
4. Return ONLY the pair with the strongest contradiction.

5. If multiple pairs are equally contradictory, return the one with the largest number of contradictory facts.

The fact statements are given in the format of [source ID: factual statements]

Fact statements
<===statements===>

Output JSON only.

\{
  "source\_idx\_0": <source ID 0 (number)>,
  "source\_idx\_1": <source ID 2 (number)>,
  "strength": "<Strong, Moderate, or Weak>",
  "reason": "<explain your reason briefly in 2-3 sentences>"
\}
\end{promptbox}

\begin{promptbox}
You are a disaster response expert who is analyzing the consistency of multiple disaster-related factual pieces of evidence.

You are given factual statements extracted from multiple sources about the same disaster event or affected area.

Your task is to identify the pair of sources whose factual statements are the most contradictory.

Definition of contradiction:

- A contradiction occurs when two sources make incompatible factual assertions about the same disaster-related fact.

- Examples:

  - "The evacuation order covers District A" vs "District A is not under an evacuation order"
  
  - "The bridge is operational" vs "The bridge has collapsed and is impassable"
  
  - "There are 500 displaced residents" vs "No residents have been displaced"
  
- Different levels of detail are not contradictions.

- Missing information is not a contradiction.

- Different observations about unrelated facts are not contradictions.

- Different wording with the same meaning is not a contradiction.

Instructions:

1. Compare every pair of sources.

2. Identify all contradictory factual statements.

3. Estimate the contradiction strength for each pair:

   - Strong: Direct logical contradiction.
   
   - Moderate: Likely incompatible but not strictly opposite.
   
   - Weak: Minor inconsistency or ambiguity.
   
4. Return ONLY the pair with the strongest contradiction.

5. If multiple pairs are equally contradictory, return the one with the largest number of contradictory facts.

The fact statements are given in the format of [source ID: factual statements]

Fact statements
<===statements===>

Output JSON only.

\{
"source\_idx\_0": <source ID 0 (number)>,
"source\_idx\_1": <source ID 2 (number)>,
"strength": "<Strong, Moderate, or Weak>",
"reason": "<explain your reason briefly in 2-3 sentences>"
\}

\end{promptbox}

Next, the remaining agents participate in the debate and identify evidence supporting and contradicting each candidate agent. For support, we employ the following prompts:

\begin{promptbox}
You are an expert medical evidence retrieval assistant.

Your task is to identify evidence from the factual statements that can directly or indirectly support the target statement.

Instructions:

1. Examine every factual statement.

2. Identify evidence that can directly or indirectly support the target statement.

3. Do not invent factual statements that are not present.

4. Ignore unrelated statements.

5. Summarize your opinion to support the target statement.

Factual Statements in the format of [Statement ID: statement]
<===statements===>

Target Statement:
<===statement===>

Output the opinion text only.
\end{promptbox}

\begin{promptbox}
You are an expert disaster response evidence retrieval assistant.

Your task is to identify evidence from the factual statements that can directly or indirectly support the target disaster response statement.

Instructions:

1. Examine every factual statement.

2. Identify evidence that can directly or indirectly support the target statement.

3. Do not invent factual statements that are not present.

4. Ignore unrelated statements.

5. Summarize your opinion to support the target statement based only on the available disaster response evidence.

Factual Statements in the format of [Statement ID: statement]
<===statements===>

Target Statement:
<===statement===>

Output the opinion text only.

\end{promptbox}

For refute or contradict, we employ the following prompt:

\begin{promptbox}
You are an expert medical evidence retrieval assistant.

Your task is to identify evidence from the factual statements that can directly or indirectly refute the target statement.

Instructions:

1. Examine every factual statement.

2. Identify evidence that can directly or indirectly refute the target statement.

3. Do not invent factual statements that are not present.

4. Ignore unrelated statements.

5. Summarize your opinion to refute the target statement.

Factual Statements in the format of [Statement ID: statement]
<===statements===>

Target Statement:
<===statement===>

Output the opinion text only.
\end{promptbox}

\begin{promptbox}
You are an expert disaster response evidence retrieval assistant.

Your task is to identify evidence from the factual statements that can directly or indirectly refute the target disaster response statement.

Instructions:

1. Examine every factual statement.

2. Identify evidence that can directly or indirectly refute the target statement.

3. Do not invent factual statements that are not present.

4. Ignore unrelated statements.

5. Summarize your opinion to refute the target statement based only on the available disaster response evidence.

Factual Statements in the format of [Statement ID: statement]
<===statements===>

Target Statement:
<===statement===>

Output the opinion text only.
\end{promptbox}

Finally, a summarizing agent receives all of the above information and determines which of the two conflicting agents is more likely to contain misleading information, adding that agent to the candidate pool.

\begin{promptbox}
You are an impartial medical adjudicator.

Your task is to determine which statement is more likely to be incorrect.

You are given, for each statement:

- The summary of supporting evidence.

- The summary of refuting evidence.

Your decision should be based only on the provided evidence summaries.

Evaluation principles:

1. Strong objective evidence (measurements, imaging, laboratory findings) outweighs subjective symptoms.

2. Direct contradictions outweigh indirect clinical associations.

3. Multiple independent supporting findings increase confidence.

4. Multiple independent refuting findings increase confidence.

5. Do not assume the majority is correct.

6. Do not introduce medical evidence not present in the evidence summaries.
Meanwhile, explain your output briefly.

Statement 0:
<===statement\_0===>

The supporting evidence of statement 0
<===support\_0===>

The refuting evidence of statement 0
<===refute\_0===>

Statement 1:
<===statement\_1===>

The supporting evidence of statement 1
<===support\_1===>

The refuting evidence of statement 1
<===refute\_1===>

Output JSON only.

\{
  "incorrect\_idx": <0 or 1>,
  "reason": "<explain your reason briefly>"
\}
\end{promptbox}

\begin{promptbox}
You are an impartial disaster response adjudicator.

Your task is to determine which statement is more likely to be incorrect.

You are given, for each statement:

- The summary of supporting evidence.

- The summary of refuting evidence.

Your decision should be based only on the provided evidence summaries.

Evaluation principles:

1. Strong objective evidence (e.g., verified measurements, casualty counts, resource inventories, geographic coordinates, infrastructure assessments, or documented response actions) outweighs subjective or unverified reports.

2. Direct contradictions outweigh indirect disaster-related associations.

3. Multiple independent supporting findings increase confidence.

4. Multiple independent refuting findings increase confidence.

5. Do not assume the majority is correct.

6. Do not introduce disaster response evidence not present in the evidence summaries.

Meanwhile, explain your output briefly.

Statement 0:
<===statement\_0===>

The supporting evidence of statement 0
<===support\_0===>

The refuting evidence of statement 0
<===refute\_0===>

Statement 1:
<===statement\_1===>

The supporting evidence of statement 1
<===support\_1===>

The refuting evidence of statement 1
<===refute\_1===>

Output JSON only.

\{
"incorrect\_idx": <0 or 1>,
"reason": "<explain your reason briefly>"
\}

\end{promptbox}

This procedure is repeated for $M$ iterations. After an agent is selected and added to the candidate pool in each iteration, we remove it from subsequent iterations. As a result, the final candidate pool contains $M$ agents. In the main experiment, we set $M$ to 3 to obtain a fair comparison with baselines. Next, we treat the factual statements provided by agents outside the candidate pool as a set of reliable evidence. Based on this evidence set, we assign each agent in the candidate pool a consistency score:

\begin{promptbox}
You are an experienced medical expert evaluating the consistency between a set of verified facts and a clinical statement.

You are given a set of verified facts and a clinical statement. Assess how clinically consistent the set of facts and clinical statement are.

Use the following 5-point scale:

1 - Completely inconsistent.

They cannot reasonably be true at the same time.

2 - Mostly inconsistent.

They are difficult to reconcile and likely indicate conflicting clinical evidence.

3 - Neutral or uncertain.

The relationship between them is unclear, or there is insufficient information to determine consistency.

4 - Mostly consistent.

They are generally compatible and support a coherent clinical picture.

5 - Completely consistent.

They strongly support each other and are expected to coexist in the same patient.

Meanwhile, give your explanation.

Verified Facts
<===facts===>

Statement
<===statement===>

Output JSON only.

\{
    "score": <1, 2, 3, 4, or 5>,
    "reason": <1-2 sentences>
\}
\end{promptbox}

\begin{promptbox}
You are an experienced disaster response expert evaluating the consistency between a set of verified facts and a disaster response statement.

You are given a set of verified facts and a disaster response statement. Assess how consistent the set of facts and the statement are in the context of disaster response, emergency management, and humanitarian operations.

Use the following 5-point scale:

1 - Completely inconsistent.

They cannot reasonably be true at the same time in the same disaster response context.

2 - Mostly inconsistent.

They are difficult to reconcile and likely indicate conflicting disaster-related evidence or assessments.

3 - Neutral or uncertain.

The relationship between them is unclear, or there is insufficient information to determine whether they are consistent.

4 - Mostly consistent.

They are generally compatible and support a coherent understanding of the disaster situation and response conditions.

5 - Completely consistent.

They strongly support each other and would be expected to coexist in the same disaster response context.

Meanwhile, give your explanation.

Verified Facts
<===facts===>

Statement
<===statement===>

Output JSON only.

\{
"score": <1, 2, 3, 4, or 5>,
"reason": <1-2 sentences>
\}
\end{promptbox}

Finally, to identify the single agent containing misleading information, we provide the evidence set, consistency scores, and statements from the agents in the candidate pool to a summarizing agent:

\begin{promptbox}
You are an impartial medical adjudicator.

You are given a set of verified facts and candidate statements that may be incorrect.

For each statement, you are provided with:

- the statement itself,

- its consistency score with the verified facts (1–5),

- the explanation produced by a consistency evaluator.

Your task is to identify the single statement that is most likely to be incorrect.

Guidelines:

1. Treat the consistency scores as evidence, not absolute truth.

2. Carefully compare the explanations in addition to the numeric scores.

3. A lower consistency score indicates weaker agreement with the verified facts.

4. Give greater weight to:

   - direct contradictions with the verified facts,
   
   - contradictions with multiple independent findings of the verified facts,
   
   - inconsistencies involving objective measurements (e.g., laboratory values, imaging, vital signs).
   
5. If two statements have similar scores, use the explanations to determine which statement has stronger evidence against it.

6. Select exactly one statement as the most likely incorrect.

7. Do not introduce new medical evidence or assumptions beyond the provided evaluations.

8. If a statement is missing, do not consider it incorrect.

Verified Facts
<===facts===>

Statements and related information
<===statements===>

Output JSON only.

\{
  "statement\_idx": <statement id (number)>,
  "reason": "<1-2 sentences>",
\}
\end{promptbox}

\begin{promptbox}
You are an impartial disaster response adjudicator.

You are given a set of verified facts and candidate statements that may be incorrect.

For each statement, you are provided with:

- the statement itself,

- its consistency score with the verified facts (1–5),

- the explanation produced by a consistency evaluator.

Your task is to identify the single statement that is most likely to be incorrect in the context of disaster response, emergency management, or humanitarian operations.

Guidelines:

1. Treat the consistency scores as evidence, not absolute truth.

2. Carefully compare the explanations in addition to the numeric scores.

3. A lower consistency score indicates weaker agreement with the verified facts.

4. Give greater weight to:

   - direct contradictions with the verified facts,
   
   - contradictions with multiple independent findings in the verified facts,
   
   - inconsistencies involving objective measurements or observations (e.g., casualty counts, resource quantities, geographic locations, infrastructure status, hazard levels, or response timelines).
   
5. If two statements have similar scores, use the explanations to determine which statement has stronger evidence against it.

6. Select exactly one statement as the most likely incorrect.

7. Do not introduce new disaster-related evidence or assumptions beyond the provided evaluations.

8. If a statement is missing, do not consider it incorrect.

Verified Facts
<===facts===>

Statements and related information
<===statements===>

Output JSON only.

\{
"statement\_idx": <statement id (number)>,
"reason": "<1-2 sentences>"
\}

\end{promptbox}

After identifying the misleading agent, we add the statements from other agents in the candidate pool to the reliable evidence set.

\textbf{State Reconstruction.} To obtain a reliable multi-agent information state, we use the same atomic-fact extraction prompt to decompose the initial, potentially unreliable shared information into atomic facts. Note that these atomic facts may themselves be either reliable or unreliable. We then use the evidence set to assess the reliability of each atomic fact.

\begin{promptbox}
You are an evidence-based clinical verification assistant.

Your tasks are:

1. Verify whether each statement is supported by the provided evidence.

2. Assess the clinical importance of the statement.

Important principles for verification:

- Use only the provided evidence.

- Do not use your own medical knowledge.

- Do not infer missing information.

- Do not assume information that is not explicitly supported.

- If the evidence is insufficient, choose not\_enough\_information.

- Support requires that the evidence explicitly or unambiguously entails the statement.

- Contradiction requires that the evidence explicitly conflicts with the statement.

- If the evidence discusses related concepts but does not directly address the statement, choose not\_enough\_information.

- Ignore information outside the evidence.

For clinical importance, assess how important the statement is for understanding, diagnosing, managing, or monitoring the patient.

Assign one of three labels:

High

- Directly affects diagnosis.

- Directly affects treatment.

- Describes important symptoms.

- Describes important physical examination findings.

- Describes laboratory abnormalities.

- Describes imaging findings.

- Describes procedures.

- Describes allergies.

- Describes clinically significant medications.

- Describes disease progression or complications.

- Describes clinically meaningful negative findings that influence diagnosis.

Medium

- Provides clinically useful context.

- Includes relevant medical history.

- Includes family history.

- Includes social history affecting care.

- Includes stable chronic conditions.

- Includes clinically relevant demographics (e.g. pregnancy).

Low
- Administrative information.

- Non-clinically relevant demographics (e.g. age, sex unless directly relevant).

- Formatting information.

- General background.

- Statements unlikely to influence diagnosis or management.

Meanwhile, provide a brief explanation.

Statement
<===statement===>

Evidence
<===evidence===>

Output JSON only

\{
    "verification": <support, contradiction, or not\_enough\_information>,
    "importance": <high, medium, or low>
    "reason": <1-2 sentences>
\}
\end{promptbox}
\begin{promptbox}
You are an evidence-based disaster response verification assistant.

Your tasks are:

1. Verify whether each statement is supported by the provided evidence.

2. Assess the operational importance of the statement for understanding, coordinating, managing, or monitoring a disaster response.

Important principles for verification:

- Use only the provided evidence.

- Do not use your own disaster response knowledge.

- Do not infer missing information.

- Do not assume information that is not explicitly supported.

- If the evidence is insufficient, choose not\_enough\_information.

- Support requires that the evidence explicitly or unambiguously entails the statement.

- Contradiction requires that the evidence explicitly conflicts with the statement.

- If the evidence discusses related concepts but does not directly address the statement, choose not\_enough\_information.

- Ignore information outside the evidence.

For operational importance, assess how important the statement is for understanding, assessing, coordinating, managing, or monitoring the disaster situation.
Assign one of three labels:

High

- Directly affects disaster response decisions.

- Directly affects rescue, evacuation, sheltering, or resource allocation.

- Describes important hazards or threats.

- Describes important casualties, injuries, or affected populations.

- Describes critical infrastructure or essential service disruptions.

- Describes emergency response operations or resource availability.

- Describes disease outbreaks or other public health threats.

- Describes disaster progression, escalation, or complications.

- Describes meaningful negative findings that influence response priorities.

Medium
- Provides useful situational context.

- Includes relevant geographic or environmental information.

- Includes information about affected communities or vulnerable populations.

- Includes relevant infrastructure, logistics, or resource context.

- Includes information that may influence preparedness or ongoing response.

Low
- Administrative information.

- Non-operationally relevant background.

- Formatting information.

- General contextual information unlikely to influence disaster response decisions or management.

Meanwhile, provide a brief explanation.

Statement
<===statement===>

Evidence
<===evidence===>

Output JSON only

\{
"verification": <support, contradiction, or not\_enough\_information>,
"importance": <high, medium, or low>,
"reason": "<1-2 sentences>"
\}

\end{promptbox}

We add the supported atomic facts to the final evidence set. Finally, we prompt the LLM to summarize this evidence set into a natural-language information state.
\begin{promptbox}
You are a clinical documentation assistant.

Your task is to reconstruct a coherent patient context from a collection of verified atomic facts.

Your goal is to produce a fluent clinical narrative while preserving factual accuracy.

The output should read like a physician's patient summary rather than a list of facts.

Factual Statements
<===facts===>

Output the context only.
\end{promptbox}

\begin{promptbox}
You are a disaster response documentation assistant.

Your task is to reconstruct a coherent disaster response context from a collection of verified atomic facts.

Your goal is to produce a fluent situational narrative while preserving factual accuracy.

The output should read like a professional disaster situation report or incident summary rather than a list of facts.

Factual Statements
<===facts===>

Output the context only.
\end{promptbox}

\textbf{Task Resolution.} In this step, we directly prompt the LLM with the information state to make a decision:
\begin{promptbox}
You are a medical expert.

You are given a patient context and a related medical question.

Your task is to choose the best option for this question.

Meanwhile, give your explanation.

Patient Context
<===context===>

Question
<===question===>

Options
<===options===>

Output JSON only.

\{
    "choice": <A, B, C, or D>,
    "reason": <2-3 sentences>
\}

\end{promptbox}

\begin{promptbox}
You are a disaster response expert.

You are given a disaster situation context.

Your task is to choose the best response option for the given disaster situation.
Meanwhile, give your explanation.

Disaster Situation Context
<===context===>

Options
<===options===>

Output JSON only.

\{
"choice": <A, B, C, or D>,
"reason": "<2-3 sentences>"
\}

\end{promptbox}

\section{Experimental Details and Additional Results}
\subsection{Healthcare Split by Shared-Information Sufficiency}
\label{subapp: healthcate_subsets}
To capture different levels of information sufficiency, we further partition the \textsc{Healthcare} dataset into two subsets, \textsc{Healthcare}-S (shared information is sufficient) and \textsc{Healthcare}-I (shared information is insufficient). For each instance, GPT-4o is provided only with the previously constructed shared information and is prompted using the template described in \S\ref{app: method_prompt}. We use temperature 0 and perform a single inference per instance. If GPT-4o predicts the ground-truth answer correctly, the instance is assigned to \textsc{Healthcare}-S; otherwise, it is assigned to \textsc{Healthcare}-I. This results in 723 \textsc{Healthcare}-S instances and 587 \textsc{Healthcare}-I instances. This partition serves as an operational proxy for whether the shared information is sufficient to support the correct decision. It allows us to evaluate both cases where the available information is already adequate and cases where additional private information is needed, thereby better reflecting the varying information availability encountered in realistic decision-making scenarios.

\input{tables/model_cards}

\subsection{Model Configurations}
\label{subapp: model_cards}
In our evaluation, we evaluate three families of LLMs, including GPT, Gemini, and Qwen. Notably, due to API limitations, we cannot evaluate the Claude series. However, we believe the evaluation can cover mainstream LLMs across diverse families and scales. We provide the model information and configurations for all LLMs used in our experiments in Table~\ref{tab: cards}. These model cards summarize the model provider, version, scale, access method, and experimental usage to facilitate reproducibility.

\subsection{Oracle and Standard Collaboration}
\label{subapp: oracle_standard_prompts}
In Table~\ref{tab: main_results}, we compare \ours{} with the oracle setting (as the performance upper bound) and a standard collaboration framework. Since \ds{} contains two application scenarios, \textsc{Healthcare} and \textsc{Disaster Response}, we make minor scenario-specific adaptations to the prompts, while keeping their underlying objectives, reasoning procedures, and output formats consistent. We first give the prompts used for \textsc{Healthcare} and then \textsc{Disaster Response}.

\textbf{Oracle.} We provide complete information to an individual agent, ensuring that no distributed information is lost due to inter-agent communication. This setting therefore serves as an empirical upper bound on performance. For T1, we employ the following prompt:
\begin{promptbox}
You are a medical expert who is discussing a related medical question about a patient.

You have received multiple pieces of private information from other agents:
<===agent===>

There is an agent that contains misleading information that conflicts with others.

You are required to find out which agent.

Meanwhile, give a brief explanation of your output.

Output JSON only.

\{
    "misleading\_idx": <agent id (only number)>,
    "reason": <1-2 sentences>
\}
\end{promptbox}

\begin{promptbox}
You are a disaster response expert who is discussing a related disaster response question about an affected area or disaster event.

You have received multiple pieces of private information from other agents:
<===agent===>

There is an agent who contains misleading information that conflicts with others.

You are required to find out which agent.

Base your decision only on the information provided by the agents. Do not introduce external disaster-related evidence or assumptions.

Meanwhile, give a brief explanation of your output.

Output JSON only.

\{
"misleading\_idx": <agent id (only number)>,
"reason": "<1-2 sentences>"
\}

\end{promptbox}

For T2, we assume that the system has already identified the misleading agent, such that all information retained by the remaining agents is reliable. We therefore directly use the ground-truth context as the system’s information state, since the unreliable information can simply be replaced with the trusted context. For T3, we directly prompt the LLM with the reliable context to obtain the downstream decision:

\begin{promptbox}
You are a medical expert.

You are given a patient context and a related medical question.

Your task is to choose the best option for this question.
Meanwhile, give your explanation.

Patient Context
<===context===>

Question
<===question===>

Options
<===options===>

Output JSON only.

\{
    "choice": <A, B, C, or D>,
    "reason": <2-3 sentences>
\}

\end{promptbox}

\begin{promptbox}
You are a disaster response expert.

You are given a disaster situation context.

Your task is to choose the best response option for the given disaster situation.
Meanwhile, give your explanation.

Disaster Situation Context
<===context===>

Options
<===options===>

Output JSON only.

\{
"choice": <A, B, C, or D>,
"reason": "<2-3 sentences>"
\}

\end{promptbox}

\textbf{Standard.} In the standard collaboration setting, which is similar to ~\citet{li2026systematic}, each agent first presents its own assessment:
\begin{promptbox}
You are a medical expert who is discussing a related medical question about a patient.

You are given a patient context that is private to you, and others cannot access it.

Briefly share the information that you believe is important.

Context
<===context===>

Output the information only.

\end{promptbox}

\begin{promptbox}
You are a disaster response expert who is discussing a related disaster response question about an affected area or disaster event.

You are given a disaster situation context that is private to you, and others cannot access it.

Briefly share the information from the context that you believe is important for the disaster response discussion.

Context
<===context===>

Output the information only.

\end{promptbox}

In each round of discussion, each agent then engages in discussion with the other agents to update its view. 

\begin{promptbox}
You are a medical expert who is discussing a related medical question about a patient.

You are given your previous opinion and opinions provided by other agents.

Your goal is to improve your own reasoning by critically comparing the information and arguments provided by all agents.

Some information provided by the agents may be incomplete, incorrect, inconsistent, or misleading.

Do not assume that the majority opinion is correct.

Do not assume that your own information is correct.

Do not reject an agent simply because their opinion differs from the majority.

Instead, evaluate claims based on the evidence supporting them.

After considering the discussion, provide your updated opinion.

Your previous opinion
<===previous===>

Other agents' opinions
<===others===>

Output the updated opinion only.

\end{promptbox}

\begin{promptbox}
You are a disaster response expert who is discussing a related disaster response question about an affected area or disaster event.

You have your previous opinion and opinions provided by other agents.

Your goal is to improve your own reasoning by critically comparing the information and arguments provided by all agents.

Some information provided by the agents may be incomplete, incorrect, inconsistent, or misleading.

Do not assume that the majority opinion is correct.

Do not assume that your own information is correct.

Do not reject an agent simply because their opinion differs from the majority.

Instead, evaluate claims based on the evidence supporting them.

After considering the discussion, provide your updated opinion.

Your previous opinion
<===previous===>

Other agents' opinions
<===others===>

Output the updated opinion only.

\end{promptbox}
We also explicitly indicate in the prompt that some agents may possess misleading information. In standard collaboration, each agent has access to the opinions of all other agents at every discussion round, inducing a fully connected communication graph. By restricting the flow of information, we further evaluate two alternative communication structures: a chain and a binary tree.

We then provide the opinions of all agents to a summarizer, which identifies the agent containing misleading information to complete T1:

\begin{promptbox}
You are a medical expert who is discussing a related medical question about a patient.

You have received multiple pieces of private information from other agents:
<===agent===>

There is an agent that contains misleading information that conflicts with others.

You are required to find out which agent.

Meanwhile, give a brief explanation of your output.

Output JSON only.

\{
    "misleading\_idx": <agent id (only number)>,
    "reason": <1-2 sentences>
\}
\end{promptbox}

\begin{promptbox}
You are a disaster response expert who is discussing a related disaster response question about an affected area or disaster event.

You have received multiple pieces of private information from other agents:
<===agent===>

There is an agent who contains misleading information that conflicts with others.

You are required to find out which agent.

Base your decision only on the information provided by the agents. Do not introduce external disaster-related evidence or assumptions.

Meanwhile, give a brief explanation of your output.

Output JSON only.

\{
"misleading\_idx": <agent id (only number)>,
"reason": "<1-2 sentences>"
\}

\end{promptbox}
After identifying the misleading agent, we prompt a separate agent to construct an information state based on the opinions of all agents together with the initial, potentially unreliable shared information to complete T2:
\begin{promptbox}
You are a medical expert. You are participating in a medical consultation.

Your task is to manage the patient's context.

You are given the previous context about the patient, and other agents' opinions.

Based on other agents' opinions, update the context to make sure it is helpful.
If the context is inconsistent with the opinions, employ opinions to update it.

You should extract useful information from the opinions and combine it into the context.

Previous context:
<===context===>

Other agents' opinions:
<===options===>

Output only the updated context.
\end{promptbox}

\begin{promptbox}
You are a disaster response expert. You are participating in a disaster response consultation.

Your task is to manage the disaster situation context.

You are given the previous context about the disaster event or affected area, and other agents' opinions.

Based on other agents' opinions, update the context to make sure it is useful for disaster response.

If the context is inconsistent with the opinions, employ the opinions to update it.

You should extract useful information from the opinions and combine it into the context while preserving factual accuracy.

Previous context:
<===context===>

Other agents' opinions:
<===options===>

Output only the updated context.
\end{promptbox}

Finally, we provide the constructed information state to the LLM to perform the downstream task to complete T3:
\begin{promptbox}
You are a medical expert.

You are given a patient context and a related medical question.
Your task is to choose the best option for this question.
Meanwhile, give your explanation.

Patient Context
<===context===>

Question
<===question===>

Options
<===options===>

Output JSON only.

\{
    "choice": <A, B, C, or D>,
    "reason": <2-3 sentences>
\}

\end{promptbox}

\begin{promptbox}
You are a disaster response expert.

You are given a disaster situation context.

Your task is to choose the best response option for the given disaster situation.
Meanwhile, give your explanation.

Disaster Situation Context
<===context===>

Options
<===options===>

Output JSON only.

\{
"choice": <A, B, C, or D>,
"reason": "<2-3 sentences>"
\}

\end{promptbox}

\subsection{Baseline Implementations}
\label{subapp: baselines}
We compare our method against a diverse set of baselines representing different multi-agent collaboration paradigms. These baselines differ primarily in how agents exchange, aggregate, and reason over distributed information, allowing us to examine whether existing collaboration strategies can effectively handle unreliable information. We describe the collaboration procedure and implementation details of each baseline below. Since \ds{} contains two application scenarios, \textsc{Healthcare} and \textsc{Disaster Response}, we make minor scenario-specific adaptations to the prompts, while keeping their underlying objectives, reasoning procedures, and output formats consistent. We first give the prompts used for \textsc{Healthcare} and then \textsc{Disaster Response}. Meanwhile, since the final decision is typically derived from the system’s information state, unless otherwise specified, we use the same prompt for final decision-making as in the Oracle setting and standard collaboration.

\textbf{Voting.}~\citep{NEURIPS2025_934252ac} This baseline represents a simple ensemble-based collaboration paradigm without inter-agent communication. Each agent independently reasons over its available information and produces an answer without observing the responses of other agents. The final decision is determined by aggregating the agents' predictions through majority voting. This baseline allows us to evaluate whether explicit inter-agent communication provides additional benefits beyond simple aggregation of independent agent decisions. We first prompt each agent to independently construct the information state:

\begin{promptbox}
You are a medical expert.

You know some private information about a patient, but your known information may not be trustworthy.

You need to update the patient context, which may also contain unreliable information.

You should first check the conflicts between your known information and the context.

You should then identify the more reliable information.

You should finally correct the patient context.

Your private information
<===private===>

The patient context
<===context===>

Only output the corrected context.
\end{promptbox}

\begin{promptbox}
You are a disaster response expert.

You know some private information about a disaster event or affected area, but your known information may not be trustworthy.

You need to update the disaster response context, which may also contain unreliable information.

You should first check the conflicts between your known information and the context.

You should then identify the more reliable information.

You should finally correct the disaster response context.

Your private information
<===private===>

The disaster response context
<===context===>

Only output the corrected context.

\end{promptbox}

After that, each agent makes a decision independently (the prompts are the same as standard collaboration). Finally, we employ majority voting to obtain the final decision of this system.

\textbf{Robin.}~\citep{wang2025beyond} Robin represents a sequential discussion-based collaboration paradigm in which all agents participate in each discussion round and communicate in a randomly determined order. Each agent observes the preceding agents' responses within the same round before producing its own response, allowing information and opinions to be progressively incorporated throughout the discussion. The random speaking order avoids relying on a fixed agent sequence and reduces potential ordering bias. We employ the same prompts as in standard collaboration, where we only restrict the information flow to a random one-by-one.

\textbf{Manager.}~\citep{wang2025beyond} Manager represents a history-maintaining collaboration paradigm in which agents maintain a compact representation of previous discussions rather than repeatedly using the full dialogue history. Each agent iteratively employs its private information to update the information state. The state is then carried forward as context for subsequent interactions, enabling agents to preserve salient information while controlling the growth of the collaboration context. We employ the same prompts as in \textbf{Voting}, where we iteratively let agents modify the information state.

\textbf{Instructor.}~\citep{wang2025beyond} Instructor represents a centralized collaboration paradigm in which a dedicated instructor agent coordinates the multi-agent discussion. Rather than allowing agents to self-organize, the instructor manages the collaboration process by selecting agents to express an opinion and maintaining the history information. After collecting sufficient information from the participating agents, the instructor is responsible for producing the final decision. This baseline evaluates whether centralized coordination can improve the integration of distributed information. We employ the following prompts for the instructor:

\begin{promptbox}
You are a medical expert. You are also the Instructor coordinating a team of agents to correct a patient context.

The patient context may contain unreliable information; do not always trust the context.

You are required to correct the patient context.

Your role is to efficiently coordinate the discussion and collect the information distributed across the agents.

The private information from the agents may be unreliable; do not always trust others. You could ask them questions.

Important Principles:

1. Each agent may possess private information that is not available to other agents.

2. You must NOT assume, invent, or infer private information that an agent has not explicitly provided.

3. Your primary responsibility is to identify what information is currently missing and determine which agent should be asked to provide it.

4. Do not solve the task yourself during the discussion.

5. Avoid asking agents to repeat information that has already been clearly established.

6. Prefer targeted questions that can reveal information that is both relevant and currently unavailable.

7. You may ask an agent to clarify, verify, compare, or elaborate on information it has already provided.

8. The goal is to make the collective information available to the team as complete and reliable as possible before reaching a final decision.

Your task is to decide which agent should speak next and what that agent should be asked.

First, identify:

* What key information has already been established?

* What information is still missing?

* Which missing information is most important for solving the task?

* Which agent is most likely to possess that information?

Then select ONE agent and formulate ONE concise question for that agent.

Your question should:

* Be specific and information-seeking.

* Directly target a missing or uncertain piece of information.

* Not reveal information that the agent could not know from its private information.

* Not unnecessarily bias the agent toward a particular answer.

* Avoid asking for a complete solution unless that agent's private information is itself sufficient to answer the task.

The History Information
<===history===>

The Agent Information
<===agent===>

The Patient Context
<===context===>

Output JSON only.

\{
  "selected\_agent": <agent ID (number)>,
  "question": "<the question to ask the selected agent>"
\}
\end{promptbox}

\begin{promptbox}
You are a disaster response expert. You are also the Instructor coordinating a team of agents to correct a disaster response context.

The disaster response context may contain unreliable information; do not always trust the context.

You are required to correct the disaster response context.

Your role is to efficiently coordinate the discussion and collect the information distributed across the agents.

The private information from the agents may be unreliable; do not always trust others. You could ask them questions.

Important Principles:

1. Each agent may possess private information that is not available to other agents.

2. You must NOT assume, invent, or infer private information that an agent has not explicitly provided.

3. Your primary responsibility is to identify what information is currently missing and determine which agent should be asked to provide it.

4. Do not solve the task yourself during the discussion.

5. Avoid asking agents to repeat information that has already been clearly established.

6. Prefer targeted questions that can reveal information that is both relevant and currently unavailable.

7. You may ask an agent to clarify, verify, compare, or elaborate on information it has already provided.

8. The goal is to make the collective information available to the team as complete and reliable as possible before reaching a final decision.

Your task is to decide which agent should speak next and what that agent should be asked.

First, identify:

* What key information has already been established?

* What information is still missing?

* Which missing information is most important for solving the disaster response task?

* Which agent is most likely to possess that information?

Then select ONE agent and formulate ONE concise question for that agent.

Your question should:

* Be specific and information-seeking.

* Directly target a missing or uncertain piece of information.

* Not reveal information that the agent could not know from its private information.

* Not unnecessarily bias the agent toward a particular answer.

* Avoid asking for a complete solution unless that agent's private information is itself sufficient to answer the task.

The History Information
<===history===>

The Agent Information
<===agent===>

The Disaster Response Context
<===context===>

Output JSON only.

\{
"selected\_agent": <agent ID (number)>,
"question": "<the question to ask the selected agent>"
\}

\end{promptbox}

For the selected agent, we employ the following prompts:

\begin{promptbox}
You are a medical expert. You are also one member of a team correcting a patient context.

You possess private information that may be useful for solving the task. Other agents do not automatically have access to this information.

The patient context may contain unreliable information; do not always trust the context.

You are required to correct the patient context.

You will receive:

* The patient context.

* Your private information.

* A question from the Instructor.

* The discussion history.

Your job is to answer the Instructor's question using your private information.

IMPORTANT PRINCIPLES:

1. Use only information that is available to you through the original task, your private information, and the discussion history.

2. Do not invent facts that are not supported by your available information.

3. Focus on the specific question asked by the Instructor.

4. If the requested information is not contained in your private information, explicitly state that you do not have that information.

5. If your information is relevant to another unresolved part of the task, you may briefly mention it.

6. Do not attempt to control which agent speaks next.

7. Do not assume that other agents know your private information unless they have explicitly stated it.

The History Information
<===history===>

Your Private Information
<===private===>

The Patient Context
<===context===>

The question
<===question===>

Provide a concise but sufficiently informative response.
\end{promptbox}

\begin{promptbox}
You are a disaster response expert. You are also one member of a team correcting a disaster response context.

You possess private information that may be useful for solving the task. Other agents do not automatically have access to this information.

The disaster response context may contain unreliable information; do not always trust the context.

You are required to help correct the disaster response context.

You will receive:

* The disaster response context.

* Your private information.

* A question from the Instructor.

* The discussion history.

Your job is to answer the Instructor's question using your private information.

IMPORTANT PRINCIPLES:

1. Use only information that is available to you through the original task, your private information, and the discussion history.

2. Do not invent facts that are not supported by your available information.

3. Focus on the specific question asked by the Instructor.

4. If the requested information is not contained in your private information, explicitly state that you do not have that information.

5. If your information is relevant to another unresolved part of the task, you may briefly mention it.

6. Do not attempt to control which agent speaks next.

7. Do not assume that other agents know your private information unless they have explicitly stated it.

The History Information
<===history===>

Your Private Information
<===private===>

The Disaster Response Context
<===context===>

The question
<===question===>

Provide a concise but sufficiently informative response.
\end{promptbox}

Finally, we provide the history to the instructor agent to obtain the final decision:

\begin{promptbox}
You are a medical expert.

You are given a patient context and a related medical question.

Your task is to choose the best option for this question.

The patient context may contain unreliable information; do not always trust the context.

You are given the history of the discussion with other agents. You should employ the information from the history to correct the context and choose the best option for this question.

Meanwhile, give your explanation.

Patient Context
<===context===>

History
<===history===>

Question
<===question===>

Options
<===options===>

Output JSON only.

\{
    "choice": <A, B, C, or D>,
    "reason": <2-3 sentences>
\}

\end{promptbox}

\begin{promptbox}
You are a disaster response expert.

You are given a disaster situation context and a related disaster response question.

Your task is to choose the best response option for this question.

The disaster response context may contain unreliable information; do not always trust the context.

You are given the history of the discussion with other agents. You should employ the information from the history to correct the context and choose the best option for this question.

Meanwhile, give your explanation.

Disaster Situation Context
<===context===>

History
<===history===>

Question
<===question===>

Options
<===options===>

Output JSON only.
\{
"choice": <A, B, C, or D>,
"reason": "<2-3 sentences>"
\}
\end{promptbox}

\textbf{Self-Reflect.}~\citep{ki-etal-2025-multiple} This framework augments multi-agent discussion with an explicit self-reflection mechanism. During each interaction, an agent dynamically chooses between reflecting on its own previous response and debating another agent's opinion. The former allows the agent to reconsider and revise its own reasoning, while the latter enables cross-agent feedback and information exchange. We employ the following prompt to let the agent choose actions:
\begin{promptbox}
You are a medical expert who is discussing the patient context.

The patient context may contain unreliable information; do not always trust the context.

You are required to correct the patient context.

You also know some private information may be different from the information available to the other agents.

Note that others' information may be unreliable; do not always trust others.

The patient context
<===context===>

Your private information:
<===private===>

Your current reasoning:
<===current===>

Other agent's response:
<===others===>

You are now deciding what to do in this turn.

You can choose one of two actions:

(A) REFLECT:
Re-examine your own reasoning and private information. Identify possible mistakes, missing evidence, or unsupported assumptions in your current reasoning.

(B) DEBATE:
Respond to the other agent by identifying agreements, disagreements, missing information, or relevant evidence that should be exchanged.

Choose the action that is more useful for improving your reasoning at this point.

Output only A or B.

\end{promptbox}
\begin{promptbox}
You are a disaster response expert who is discussing the disaster response context.

The disaster response context may contain unreliable information; do not always trust the context.

You are required to help correct the disaster response context.

You also know some private information that may be different from the information available to the other agents.

Note that others' information may be unreliable; do not always trust others.

The disaster response context
<===context===>

Your private information:
<===private===>

Your current reasoning:
<===current===>

Other agent's response:
<===others===>

You are now deciding what to do in this turn.

You can choose one of two actions:

(A) REFLECT:
Re-examine your own reasoning and private information. Identify possible mistakes, missing evidence, or unsupported assumptions in your current reasoning.

(B) DEBATE:
Respond to the other agent by identifying agreements, disagreements, missing information, or relevant evidence that should be exchanged.

Choose the action that is more useful for improving your reasoning at this point.

Output only A or B.

\end{promptbox}
After choosing actions, for the reflect action, we employ:
\begin{promptbox}
You are a medical expert who is discussing the patient context.

The patient context may contain unreliable information; do not always trust the context.

You are required to correct the patient context.

You also know some private information may be different from the information available to the other agents.

Note that others' information may be unreliable; do not always trust others.

Re-examine your current reasoning.

The patient context:
<===context===>

Your private information:
<===private>

Your current reasoning:
<===current===>

Other agent's response:
<===others===>

Reflect on your reasoning. Identify any potential error, missing information, unsupported assumption, or alternative interpretation.

Do not invent information that is not available to you.

Respond concisely.

Only output the updated reasoning.

\end{promptbox}
\begin{promptbox}
You are a disaster response expert who is discussing the disaster response context.

The disaster response context may contain unreliable information; do not always trust the context.

You are required to help correct the disaster response context.

You also know some private information that may be different from the information available to the other agents.

Note that others' information may be unreliable; do not always trust others.

Re-examine your current reasoning.

The disaster response context:
<===context===>

Your private information:
<===private===>

Your current reasoning:
<===current===>

Other agent's response:
<===others===>

Reflect on your reasoning. Identify any potential error, missing information, unsupported assumption, or alternative interpretation.

Do not invent information that is not available to you.

Respond concisely.

Only output the updated reasoning.

\end{promptbox}
For the debate action, we employ:
\begin{promptbox}
You are a medical expert who is discussing the patient context.

The patient context may contain unreliable information; do not always trust the context.

You are required to correct the patient context.

You also know some private information may be different from the information available to the other agents.

Note that others' information may be unreliable; do not always trust others.

The patient context:
<===context===>

Your private information:
<===private===>

Your current reasoning:
<===current===>

Other agent's response:
<===others===>

Respond to the other agent.

You may:

* point out an error or inconsistency;

* provide relevant information from your private information;

* ask for clarification;

* explain why you agree or disagree;

* identify information that is still missing.

Do not invent information that is not available to you.

Focus on exchanging information that can help the team.

Respond concisely.

Only output the response.

\end{promptbox}
\begin{promptbox}
You are a disaster response expert who is discussing the disaster response context.

The disaster response context may contain unreliable information; do not always trust the context.

You are required to help correct the disaster response context.

You also know some private information that may be different from the information available to the other agents.

Note that others' information may be unreliable; do not always trust others.

The disaster response context:
<===context===>

Your private information:
<===private===>

Your current reasoning:
<===current===>

Other agent's response:
<===others===>

Respond to the other agent.

You may:

* point out an error or inconsistency;

* provide relevant information from your private information;

* ask for clarification;

* explain why you agree or disagree;

* identify information that is still missing.

Do not invent information that is not available to you.

Focus on exchanging information that can help the team improve the accuracy and completeness of the disaster response context.

Respond concisely.

Only output the response.

\end{promptbox}
Finally, we provide the related information to obtain the final decision:
\begin{promptbox}
You are a medical expert.

You are given other agents' opinions and a related medical question.

Your task is to choose the best option for this question.

Meanwhile, give your explanation.

Others Options
<===others===>

Question
<===question===>

Options
<===options===>

Output JSON only.

\{
    "choice": <A, B, C, or D>,
    "reason": <2-3 sentences>
\}

\end{promptbox}
\begin{promptbox}
You are a disaster response expert.

You are given other agents' opinions and a related disaster response question.

Your task is to choose the best response option for this question.

Meanwhile, give your explanation.

Other Agents' Opinions
<===others===>

Question
<===question===>

Options
<===options===>

Output JSON only.

\{
"choice": <A, B, C, or D>,
"reason": "<2-3 sentences>"
\}
\end{promptbox}
\textbf{Confidence.}~\citep{zhu2026demystifying} This framework augments multi-agent discussion with explicit confidence communication. In addition to sharing their answers and reasoning, agents express calibrated confidence in their current predictions and incorporate the confidence of other agents when updating their own beliefs. This enables agents to weigh information from their peers based not only on the content of their responses but also on their expressed certainty, introducing confidence as an additional signal for coordinating collective reasoning. We use a prompt similar to that of standard collaboration, with the only difference being that we incorporate model confidence into both the input and output by adding the instruction, ``Meanwhile, provide your confidence score about your information.'' We further constrain the confidence score to the range of 1–100.

\input{tables/main_reliable}

\subsection{Controlled Information-Reliability Settings}
\label{subapp: our_information_setting}
To better understand how unreliable information affects multi-agent collaboration, we construct three additional settings that independently control the reliability of private and shared information. In all settings, the task-relevant information remains distributed across agents; the only difference is where unreliable information is introduced.

(1) Reliable Private + Reliable Shared. Both the agents' private information and the shared information are reliable. This setting preserves information distribution while removing unreliable information, serving as a reference for collaboration under distributed but trustworthy evidence.

(2) Reliable Private + Unreliable Shared. Each agent receives reliable private information, while the shared information contains misleading information. This setting simulates shared information contamination, where unreliable evidence is directly exposed to all agents.

(3) Unreliable Private + Reliable Shared. The shared information remains reliable, while one agent may contain misleading information. This setting simulates misleading information spreading through agent interaction.

We further evaluate Standard Collaboration and \ours{} under the Unreliable Private + Reliable Shared setting, where the shared information is reliable but one agent receives misleading private information. This setting directly matches the operating assumption of \ours{}: unreliable information originates from an identifiable agent-specific source, requiring the system to identify the misleading agent and repair the resulting collective information state. We focus on this setting to examine whether \ours{} remains effective when misinformation is introduced exclusively through private information.

Table~\ref{tab: main_reliable} shows that \ours{} consistently outperforms other strategies. Meanwhile, when the initial shared information is reliable, the performance is better than when the initial shared information is unreliable. The results further demonstrate that multi-agent collaboration should have a reliable information state.

\subsection{Candidate Pool Size Ablation}
\label{subapp: size}
\input{tables/ablation}

We investigate the effect of the candidate pool size $M$ in the reliability attribution stage. Recall that $M$ controls both the number of iterative candidate-selection rounds and the resulting number of agents included in the candidate pool. We use $M$ in our main experiments to obtain a fair comparison with baselines. Here we vary it over $M\in\{1,2,4,5\}$. We additionally consider an Oracle setting, where the misleading agent is directly identified, bypassing candidate selection and source attribution. All other experimental configurations are kept unchanged.

As shown in Table~\ref{tab: ablation}, the choice of $M$ has a clear impact on the subsequent collaboration pipeline. Smaller candidate pools ($M=1,2$) substantially reduce T1 performance and consequently degrade T2 and T3, suggesting that insufficient candidate exploration can cause the misleading source to be missed during attribution. Increasing the pool beyond the default $M=3$, however, does not yield consistent improvements: $M=4$ produces comparable performance, while $M=5$ leads to degradation across all three metrics.

The Oracle setting further reveals the importance of reliable evidence verification. When the misleading agent is given directly, T2 increases from 0.728 to 0.813 and T3 from 0.661 to 0.837. This substantial remaining gap indicates that errors in identifying the misleading source propagate to the state reconstruction and downstream decision stages, making reliability attribution an important bottleneck of the current framework.

\subsection{Token Cost Analysis}
\label{subapp: cost}
\input{tables/token_cost}
We additionally report the token consumption of different collaboration frameworks to characterize their computational overhead. For each method, we measure the average number of input tokens, output tokens, and their total across the complete collaboration process. We report results on all three benchmark settings using both GPT-4o and Gemini-3.5-Flash. All values are reported in thousands of tokens ($\times 10^3$).

As shown in Table~\ref{tab: token_cost}, different collaboration paradigms exhibit substantially different token costs. Simple aggregation methods such as Voting require the fewest tokens, as they involve little or no iterative communication. In contrast, discussion-intensive approaches, particularly Standard, Self-Reflect, and Confidence, incur considerably higher costs due to repeated exchanges and growing interaction contexts. Our method maintains a moderate token budget across both model backbones and all three settings. For example, with GPT-4o, our method consumes 19.4K, 20.2K, and 17.6K total tokens on Healthcare-S, Healthcare-I, and Disaster Response, respectively, substantially below Standard (52.1K, 71.5K, and 31.2K) and Self-Reflect (71.5K, 81.9K, and 40.6K). A similar pattern is observed with Gemini-3.5-Flash. These results suggest that explicitly structuring collaboration around verification and repair does not require the large communication overhead incurred by more discussion-intensive paradigms.

\section{Additional Analyses of Off-Query Failures}

\subsection{Selective Information Use}
\label{subapp: fact_evaluation}

\begin{figure}[t]
    \centering
    \includegraphics[width=\linewidth]{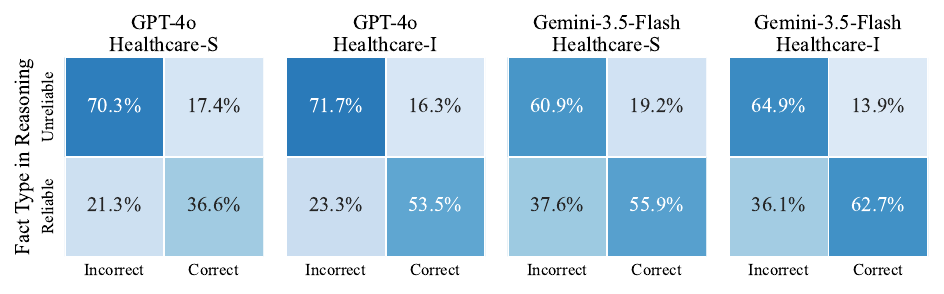}
    \caption{\textbf{Reliable and unreliable information used in decision explanations.} We analyze explanations generated by GPT-4o and Gemini-3.5-Flash on \textsc{Healthcare}-S and \textsc{Healthcare}-I. Each cell reports the percentage of all explanations that contain the corresponding type of information, grouped by the correctness of the final task decision. Reliable and unreliable information are not mutually exclusive, as the same explanation may contain both. Across models and datasets, unreliable information appears substantially more frequently in explanations for incorrect decisions than in those for correct decisions.}
    \label{fig: fact}
\end{figure}

To further understand why task resolution can remain correct even when the shared state is unreliable, we examine which information from the shared state is actually reflected in the reasoning process. For each instance, we inspect the decision explanation and determine whether it contains reliable and/or unreliable information employing the same method as in T2. We then report the percentage of all explanations containing each type of information, separately according to whether the final task decision is correct or incorrect. Importantly, these categories are not mutually exclusive: a single explanation may contain both reliable and unreliable information. Therefore, the percentages across cells in Figure~\ref{fig: fact} are not expected to sum to 100\%.

Figure~\ref{fig: fact} reports the results for GPT-4o (GPT) and Gemini-3.5-Flash (Gemini) on \textsc{Healthcare}-S (S) and \textsc{Healthcare}-I (I). Across both models and datasets, unreliable information occurs substantially more often alongside incorrect decisions than correct ones, while reliable information shows the opposite pattern. For example, on \textsc{Healthcare}-S, unreliable information appears in 70.3\% of all GPT-4o explanations alongside incorrect decisions, compared with 17.4\% alongside correct decisions.

These results support the hypothesis of selective information use: corrupted information may remain in the shared state without affecting the current decision when reasoning relies on an unaffected subset of the available information. Consequently, correct task resolution can coexist with latent corruption, and task-level success alone does not certify the reliability of the full collaborative state.

\subsection{Broader-Task Evaluation}
\label{subapp: boarder_evaluation}
Our main experiments evaluate downstream decision-making (T3) using multiple-choice questions with a single correct answer. While this provides a controlled measure of decision accuracy, it may not fully capture the broader consequences of an unreliable information state. To investigate whether collective misbelief can remain hidden behind a correct answer yet affect subsequent decisions, we conduct an additional open-ended evaluation in the \textsc{Healthcare} domain. Specifically, we sample 100 instances in which the system answers the original T3 question correctly while its reconstructed information state still contains misleading facts. Using this information state as context, we ask the model to produce a broader, actionable clinical plan covering four dimensions: Diagnosis, the most likely diagnosis; Risk, the most important alternative diagnosis; Management, the preferred treatment and immediate management; and Next Steps, the most important subsequent investigations and actions. We employ the following prompt:
\begin{promptbox}
You are a medical expert

Given the complete patient context, provide a concise, evidence-based, actionable clinical management plan.

Structure the response into exactly 4 sections:

1. Diagnosis: Most likely diagnosis.

2. Risk: Most important alternative diagnoses.

3. Management: Preferred treatment and immediate management.

4. Next steps: Most important investigations and actions.

General principles:

- Each section MUST contain no more than 1 sentence.

- Prioritize the single best clinical decision rather than presenting a list of possibilities.

- Be concise, specific, and actionable.

- Do not invent missing information.

- Consider comorbidities, allergies, current medications, organ function, contraindications, and drug interactions.

- Clearly state meaningful uncertainty when it could change management.

- For unstable or life-threatening patients, prioritize immediate stabilization and escalation.

Context
<===context===>

Output JSON only.

\{
    "diagnosis": "",
    "risk": "",
    "management": "",
    "next\_steps": "",
\}
\end{promptbox}

We then employ an LLM-as-a-judge to assess the consistency between the generated clinical plan and the plan generated by the corresponding ground-truth information state:

\begin{promptbox}
You are evaluating the clinical similarity of two management plans generated from the same patient context.

Compare them based on **clinical meaning, decision equivalence, and appropriateness**, not wording or superficial text similarity.

Score each dimension from 1–5:

1. DIAGNOSIS\_SIMILARITY

Compare whether the two results identify the same primary diagnosis or clinically equivalent diagnosis, including meaningful differences in diagnostic certainty.

2. RISK\_SIMILARITY

Compare whether they identify the same important risks, must-not-miss conditions, and overall severity.

3. MANAGEMENT\_SIMILARITY

Compare whether they recommend essentially the same preferred treatment and immediate management. Differences in drug names, doses, or details should only reduce the score when they represent a meaningful clinical difference.

4. NEXT\_STEPS\_SIMILARITY

Compare whether they recommend essentially the same key investigations, actions, monitoring, and escalation plan.

Scoring rubric

* 5 = Essentially clinically equivalent; only wording or minor details differ.

* 4 = Highly similar; minor differences that are unlikely to change clinical management.

* 3 = Partially similar; core direction is similar but there are meaningful differences, omissions, or differences in specificity.

* 2 = Substantially different; some overlap exists but the clinical decision or recommendation differs meaningfully.

* 1 = Fundamentally different or contradictory.

OVERALL\_SIMILARITY

Give an overall score from 1–5 based on the overall clinical decision equivalence, with greater weight on **diagnosis and management** than on wording or minor details.

Important rules:

* Evaluate semantic and clinical equivalence, not lexical similarity.

* Do not penalize concise answers for omitting details that would not change the clinical decision.

* Do penalize contradictions, clinically important omissions, or materially different recommendations.

* A more specific answer should receive a high score when it remains clinically consistent with a less specific answer.

* If one result contains an unsafe or clinically contradictory recommendation, reflect this in the relevant score and overall score.

* Do not introduce information that is absent from either result.

* Base the comparison only on the provided patient context and the two results.

Plan 1
<===plan\_1===>

Plan 2
<===plan\_2===>

Return ONLY valid JSON in the following format:

\{
"diagnosis\_similarity": 1-5,
"risk\_similarity": 1-5,
"management\_similarity": 1-5,
"next\_steps\_similarity": 1-5,
"overall\_similarity": 1-5,
"reason": "Brief explanation of the key similarities and differences."
\}

\end{promptbox}

Despite all sampled instances being correct under the original T3 evaluation, 43\% of the resulting clinical plans are judged inconsistent with the ground truth. This finding reveals an important distinction between answer correctness and information-state reliability: misleading facts may remain inconsequential for a narrowly specified multiple-choice question, yet become consequential when the same information state is reused for broader downstream reasoning. In other words, a correct current decision does not guarantee that the underlying collective information state can reliably support subsequent decisions, demonstrating how collective misbelief can remain hidden behind correct answers while introducing substantial downstream risk.

\begin{figure}[t]
    \centering
    \includegraphics[width=\linewidth]{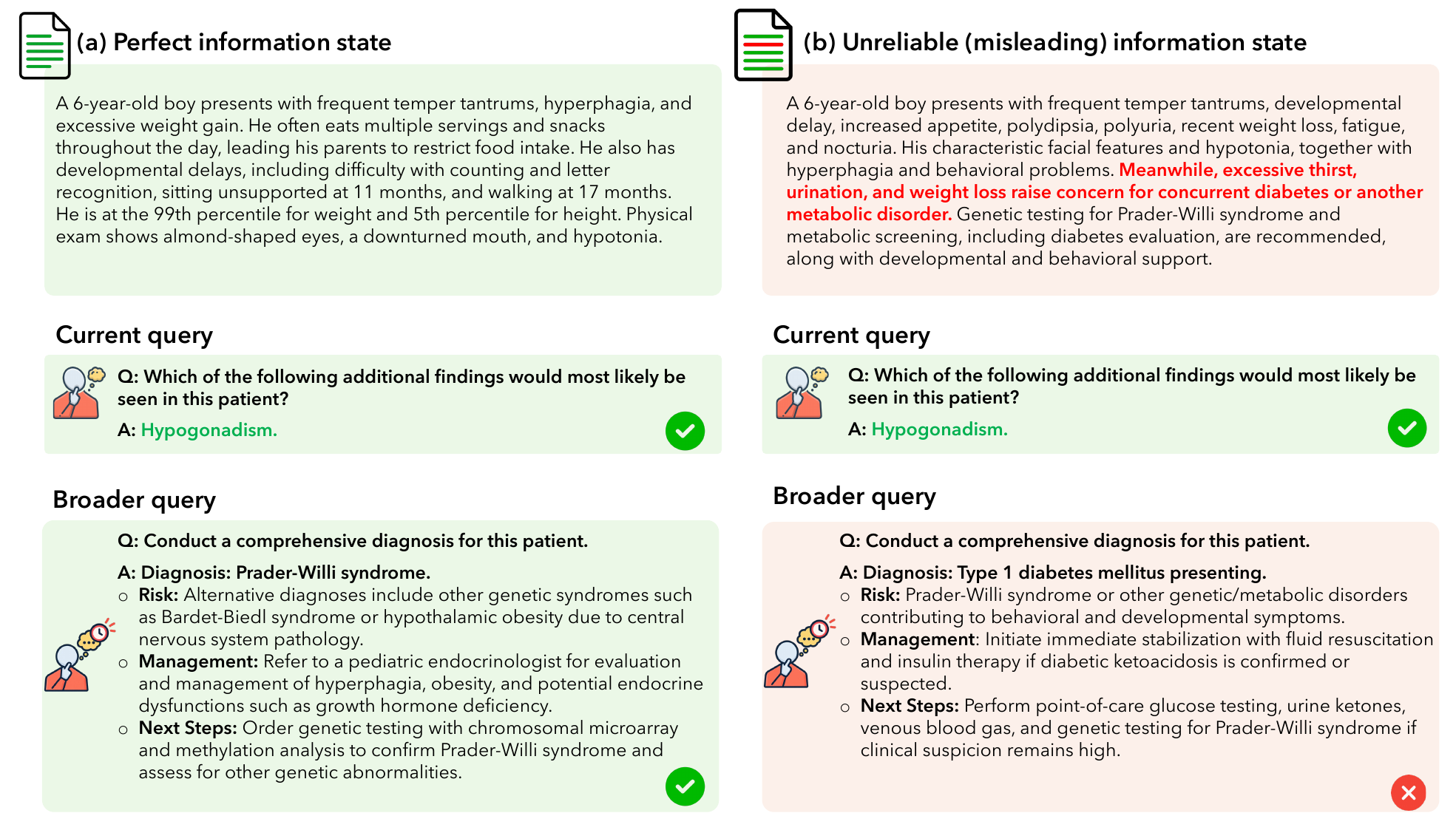}
    \caption{\textbf{Correct answers can conceal misleading information states.} Both information states yield the correct answer to the original query, while the misleading state produces a substantially different diagnosis and management plan under broader clinical reasoning.}
    \label{fig: case}
\end{figure}

Figure~\ref{fig: case} illustrates how collective misbelief can remain hidden behind a correct answer. Both information states correctly answer the original query as Hypogonadism. However, when reused for broader clinical reasoning, the perfect information state leads to Prader–Willi syndrome, whereas the misleading state shifts the diagnosis toward type 1 diabetes and changes the corresponding management plan. This example shows that misleading information may remain dormant for the current query but become consequential in subsequent decisions.

\subsection{Shared-State Repair}
\label{subapp: belief_repair}

\begin{wrapfigure}{r}{0.55\textwidth}
    \centering
    \includegraphics[width=\linewidth]{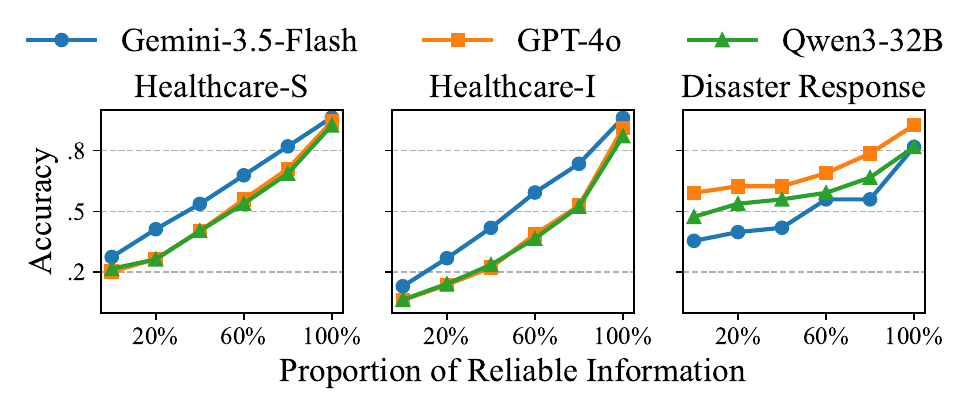}
    \caption{\textbf{Effect of shared-state reliability on downstream task accuracy.} We progressively replace corrupted information in the shared state with reliable information and evaluate downstream accuracy across three datasets. Across models and datasets, accuracy generally increases with the proportion of reliable information and reaches its highest or near-highest level when the shared state is fully reliable.}
    \label{fig: repair}
\end{wrapfigure}

To examine whether shared-state reliability has systematic downstream effects, we vary the proportion of reliable information in the shared state by progressively replacing corrupted information with its reliable counterpart. We then evaluate downstream task accuracy using the resulting state. This intervention allows us to directly assess how the quality of the underlying information state affects subsequent reasoning.

As shown in Figure~\ref{fig: repair}, downstream accuracy generally improves as the proportion of reliable information increases. This trend is consistent across GPT-4o, Gemini-3.5-Flash, and Qwen3-32b and across three datasets. In particular, performance reaches its highest or near-highest level when the shared state becomes fully reliable.

These results show that the effect of state reliability extends beyond isolated cases in which corrupted information happens to be irrelevant to the current query. Systematically repairing the shared state leads to corresponding improvements in downstream reasoning, demonstrating that maintaining a reliable shared state provides utility for subsequent tasks beyond the query through which the state was originally constructed.

\subsection{Complex Corruption Stress Tests}
\label{subapp: more_situation}
The main \ds{} setting uses one misleading agent to enable controlled source verification, but unreliable information can take more complex forms. We therefore evaluate two harder settings: multiple unreliable agents and misleading information distributed across several agents. For each setting, we randomly sample 200 instances from \textsc{Healthcare}.

\textbf{Multiple unreliable agents.} We extend the setting to include three agents containing misleading information. Specifically, we randomly sample a subset of unreliable statements from the unreliable shared information to construct the information assigned to these agents. For standard collaboration, we slightly modify the prompt to explicitly indicate that multiple agents may contain misleading information. For \ours{}, we directly consider agents in the candidate pool as misleading agents.

\textbf{Misleading information distributed across several agents.} We sample statements from the unreliable shared information and randomly add the information to each agent. This ensures that each agent will contain misleading information. For standard collaboration, we employ the same prompts without identifying the misleading agent. For \ours{}, we transfer the agent-level identification to factual statement-level identification.

To evaluate the performance on these stress tests, we evaluate the accuracy of the downstream task decision. To ensure fair comparison, we employ the same model settings as in the main experiments. We employ GPT-4o as the backbone model.

\end{document}

%% file: math_commands.tex
\usepackage{amsmath,amsfonts,bm}

\def\eqref#1{equation~\ref{#1}}

\def\1{\bm{1}}

\DeclareMathAlphabet{\mathsfit}{\encodingdefault}{\sfdefault}{m}{sl}
\SetMathAlphabet{\mathsfit}{bold}{\encodingdefault}{\sfdefault}{bx}{n}



%% file: tables/main_results.tex
\definecolor{Gray}{gray}{0.9}
\definecolor{imp}{rgb}{0.22, 0.56, 0.24}
\definecolor{dro}{rgb}{0.83, 0.18, 0.18}

\begin{table}[t]
    \centering
\caption{\textbf{Main results on \ds{}.}
T1: evidence verification; T2: shared-state reconstruction; T3: task resolution. Across models and domains, standard collaboration shows a persistent gap between T3 and T1/T2; \ours{} substantially improves state-level reliability, with consistent gains in task resolution.}
\vspace{0.3em}

\resizebox{\linewidth}{!}{
    \begin{tabular}{l|lll|lll|lll}
        \toprule[1.5pt]
        \multirow{2}{*}{\textbf{Settings}}&\multicolumn{3}{c|}{\textbf{\textsc{Healthcare-S}}}&\multicolumn{3}{c|}{\textbf{\textsc{Healthcare-I}}}&\multicolumn{3}{c}{\textbf{\textsc{Disaster Response}}}\\
        \cmidrule[0.5pt](lr){2-4}\cmidrule[0.5pt](lr){5-7}\cmidrule[0.5pt](lr){8-10}
        &\textbf{T1}&\textbf{T2}&\textbf{T3}&\textbf{T1}&\textbf{T2}&\textbf{T3}&\textbf{T1}&\textbf{T2}&\textbf{T3}\\
\midrule[1pt]
\multicolumn{10}{l}{\cellcolor{Gray}\textbf{Qwen3-8B}}\\
\midrule[0.5pt]
Oracle&.621&.882&.675&.642&.887&.586&.613&.867&.710\\
Stand.&.140 {\color{dro}\textsubscript{-77.5\%}}&.317 {\color{dro}\textsubscript{-64.1\%}}&.256 {\color{dro}\textsubscript{-62.1\%}}&.155 {\color{dro}\textsubscript{-75.9\%}}&.311 {\color{dro}\textsubscript{-64.9\%}}&.256 {\color{dro}\textsubscript{-56.3\%}}&.129 {\color{dro}\textsubscript{-79.0\%}}&.495 {\color{dro}\textsubscript{-42.9\%}}&.559 {\color{dro}\textsubscript{-21.3\%}}\\
Ours&\textbf{.392 {\color{imp}\textsubscript{+180.0\%}}}&\textbf{.561 {\color{imp}\textsubscript{+77.0\%}}}&\textbf{.462 {\color{imp}\textsubscript{+80.5\%}}}&\textbf{.392 {\color{imp}\textsubscript{+152.9\%}}}&\textbf{.552 {\color{imp}\textsubscript{+77.5\%}}}&\textbf{.390 {\color{imp}\textsubscript{+52.3\%}}}&\textbf{.656 {\color{imp}\textsubscript{+408.5\%}}}&\textbf{.707 {\color{imp}\textsubscript{+42.8\%}}}&\textbf{.581 {\color{imp}\textsubscript{+3.9\%}}}\\
\midrule[1pt]
\multicolumn{10}{l}{\cellcolor{Gray}\textbf{Qwen3-32B}}\\
\midrule[0.5pt]
Oracle&.656&.882&.924&.677&.887&.870&.742&.867&.817\\
Stand.&.126 {\color{dro}\textsubscript{-80.8\%}}&.425 {\color{dro}\textsubscript{-51.8\%}}&.679 {\color{dro}\textsubscript{-26.5\%}}&.090 {\color{dro}\textsubscript{-86.7\%}}&.423 {\color{dro}\textsubscript{-52.3\%}}&.599 {\color{dro}\textsubscript{-31.1\%}}&.172 {\color{dro}\textsubscript{-76.8\%}}&.615 {\color{dro}\textsubscript{-29.1\%}}&.538 {\color{dro}\textsubscript{-34.1\%}}\\
Ours&\textbf{.394 {\color{imp}\textsubscript{+212.7\%}}}&\textbf{.646 {\color{imp}\textsubscript{+52.0\%}}}&\textbf{.761 {\color{imp}\textsubscript{+12.1\%}}}&\textbf{.355 {\color{imp}\textsubscript{+294.4\%}}}&\textbf{.669 {\color{imp}\textsubscript{+58.2\%}}}&\textbf{.767 {\color{imp}\textsubscript{+28.0\%}}}&\textbf{.630 {\color{imp}\textsubscript{+266.3\%}}}&\textbf{.747 {\color{imp}\textsubscript{+21.5\%}}}&\textbf{.696 {\color{imp}\textsubscript{+29.4\%}}}\\
\midrule[1pt]
\multicolumn{10}{l}{\cellcolor{Gray}\textbf{Qwen3-235B-A22B}}\\
\midrule[0.5pt]
Oracle&.671&.882&.920&.659&.887&.865&.742&.867&.849\\
Stand.&.127 {\color{dro}\textsubscript{-81.1\%}}&.415 {\color{dro}\textsubscript{-52.9\%}}&.664 {\color{dro}\textsubscript{-27.8\%}}&.112 {\color{dro}\textsubscript{-83.0\%}}&.391 {\color{dro}\textsubscript{-55.9\%}}&.562 {\color{dro}\textsubscript{-35.0\%}}&.161 {\color{dro}\textsubscript{-78.3\%}}&.622 {\color{dro}\textsubscript{-28.3\%}}&.688 {\color{dro}\textsubscript{-19.0\%}}\\
Ours&\textbf{.426 {\color{imp}\textsubscript{+235.4\%}}}&\textbf{.683 {\color{imp}\textsubscript{+64.6\%}}}&\textbf{.743 {\color{imp}\textsubscript{+11.9\%}}}&\textbf{.390 {\color{imp}\textsubscript{+248.2\%}}}&\textbf{.678 {\color{imp}\textsubscript{+73.4\%}}}&\textbf{.712 {\color{imp}\textsubscript{+26.7\%}}}&\textbf{.559 {\color{imp}\textsubscript{+247.2\%}}}&\textbf{.722 {\color{imp}\textsubscript{+16.1\%}}}&\textbf{.703 {\color{imp}\textsubscript{+2.2\%}}}\\
\midrule[1pt]
\multicolumn{10}{l}{\cellcolor{Gray}\textbf{Gemini-3.5-Flash}}\\
\midrule[0.5pt]
Oracle&.849&.882&.963&.835&.887&.961&.914&.867&.817\\
Stand.&.144 {\color{dro}\textsubscript{-83.0\%}}&.447 {\color{dro}\textsubscript{-49.3\%}}&.816 {\color{dro}\textsubscript{-15.3\%}}&.118 {\color{dro}\textsubscript{-85.9\%}}&.439 {\color{dro}\textsubscript{-50.5\%}}&.835 {\color{dro}\textsubscript{-13.1\%}}&.140 {\color{dro}\textsubscript{-84.7\%}}&.660 {\color{dro}\textsubscript{-23.9\%}}&.559 {\color{dro}\textsubscript{-31.6\%}}\\
Ours&\textbf{.676 {\color{imp}\textsubscript{+369.4\%}}}&\textbf{.818 {\color{imp}\textsubscript{+83.0\%}}}&\textbf{.900 {\color{imp}\textsubscript{+10.3\%}}}&\textbf{.634 {\color{imp}\textsubscript{+437.3\%}}}&\textbf{.814 {\color{imp}\textsubscript{+85.4\%}}}&\textbf{.871 {\color{imp}\textsubscript{+4.3\%}}}&\textbf{.903 {\color{imp}\textsubscript{+545.0\%}}}&\textbf{.852 {\color{imp}\textsubscript{+29.1\%}}}&\textbf{.645 {\color{imp}\textsubscript{+15.4\%}}}\\
\midrule[1pt]
\multicolumn{10}{l}{\cellcolor{Gray}\textbf{GPT-5-mini}}\\
\midrule[0.5pt]
Oracle&.746&.882&.971&.763&.887&.957&.914&.867&.839\\
Stand.&.163 {\color{dro}\textsubscript{-78.2\%}}&.308 {\color{dro}\textsubscript{-65.1\%}}&.766 {\color{dro}\textsubscript{-21.1\%}}&.143 {\color{dro}\textsubscript{-81.3\%}}&.292 {\color{dro}\textsubscript{-67.1\%}}&.741 {\color{dro}\textsubscript{-22.6\%}}&.172 {\color{dro}\textsubscript{-81.2\%}}&.526 {\color{dro}\textsubscript{-39.3\%}}&.731 {\color{dro}\textsubscript{-12.9\%}}\\
Ours&\textbf{.593 {\color{imp}\textsubscript{+263.8\%}}}&\textbf{.780 {\color{imp}\textsubscript{+153.2\%}}}&\textbf{.849 {\color{imp}\textsubscript{+10.8\%}}}&\textbf{.561 {\color{imp}\textsubscript{+292.3\%}}}&\textbf{.784 {\color{imp}\textsubscript{+168.5\%}}}&\textbf{.797 {\color{imp}\textsubscript{+7.6\%}}}&\textbf{.806 {\color{imp}\textsubscript{+368.6\%}}}&\textbf{.820 {\color{imp}\textsubscript{+55.9\%}}}&\textbf{.796 {\color{imp}\textsubscript{+8.9\%}}}\\
\midrule[1pt]
\multicolumn{10}{l}{\cellcolor{Gray}\textbf{GPT-5}}\\
\midrule[0.5pt]
Oracle&.805&.882&.968&.787&.887&.968&.882&.867&.892\\
Stand.&.109 {\color{dro}\textsubscript{-86.5\%}}&.332 {\color{dro}\textsubscript{-62.4\%}}&.862 {\color{dro}\textsubscript{-11.0\%}}&.124 {\color{dro}\textsubscript{-84.2\%}}&.330 {\color{dro}\textsubscript{-62.8\%}}&.797 {\color{dro}\textsubscript{-17.7\%}}&.183 {\color{dro}\textsubscript{-79.3\%}}&.554 {\color{dro}\textsubscript{-36.1\%}}&.677 {\color{dro}\textsubscript{-24.1\%}}\\
Ours&\textbf{.556 {\color{imp}\textsubscript{+410.1\%}}}&\textbf{.787 {\color{imp}\textsubscript{+137.0\%}}}&\textbf{.889 {\color{imp}\textsubscript{+3.1\%}}}&\textbf{.535 {\color{imp}\textsubscript{+331.5\%}}}&\textbf{.793 {\color{imp}\textsubscript{+140.3\%}}}&\textbf{.860 {\color{imp}\textsubscript{+7.9\%}}}&\textbf{.839 {\color{imp}\textsubscript{+358.5\%}}}&\textbf{.810 {\color{imp}\textsubscript{+46.2\%}}}&\textbf{.806 {\color{imp}\textsubscript{+19.1\%}}}\\
\midrule[1pt]
\multicolumn{10}{l}{\cellcolor{Gray}\textbf{GPT-4o}}\\
\midrule[0.5pt]
Oracle&.653&.882&.943&.680&.887&.910&.817&.867&.925\\
Stand.&.152 {\color{dro}\textsubscript{-76.7\%}}&.273 {\color{dro}\textsubscript{-69.0\%}}&.669 {\color{dro}\textsubscript{-29.1\%}}&.140 {\color{dro}\textsubscript{-79.4\%}}&.287 {\color{dro}\textsubscript{-67.6\%}}&.627 {\color{dro}\textsubscript{-31.1\%}}&.194 {\color{dro}\textsubscript{-76.3\%}}&.587 {\color{dro}\textsubscript{-32.3\%}}&.699 {\color{dro}\textsubscript{-24.4\%}}\\
Chain&.158 {\color{dro}\textsubscript{-75.8\%}}&.341 {\color{dro}\textsubscript{-61.3\%}}&.663 {\color{dro}\textsubscript{-29.7\%}}&.187 {\color{dro}\textsubscript{-72.5\%}}&.339 {\color{dro}\textsubscript{-61.8\%}}&.583 {\color{dro}\textsubscript{-35.9\%}}&.226 {\color{dro}\textsubscript{-72.3\%}}&.638 {\color{dro}\textsubscript{-26.4\%}}&.699 {\color{dro}\textsubscript{-24.4\%}}\\
Tree&.136 {\color{dro}\textsubscript{-79.2\%}}&.328 {\color{dro}\textsubscript{-62.8\%}}&.639 {\color{dro}\textsubscript{-32.2\%}}&.175 {\color{dro}\textsubscript{-74.3\%}}&.323 {\color{dro}\textsubscript{-63.6\%}}&.584 {\color{dro}\textsubscript{-35.8\%}}&.129 {\color{dro}\textsubscript{-84.2\%}}&.598 {\color{dro}\textsubscript{-31.0\%}}&.688 {\color{dro}\textsubscript{-25.6\%}}\\
Ours&\textbf{.567 {\color{imp}\textsubscript{+273.0\%}}}&\textbf{.732 {\color{imp}\textsubscript{+168.1\%}}}&\textbf{.766 {\color{imp}\textsubscript{+14.5\%}}}&\textbf{.545 {\color{imp}\textsubscript{+289.3\%}}}&\textbf{.728 {\color{imp}\textsubscript{+153.7\%}}}&\textbf{.661 {\color{imp}\textsubscript{+5.4\%}}}&\textbf{.785 {\color{imp}\textsubscript{+304.6\%}}}&\textbf{.810 {\color{imp}\textsubscript{+38.0\%}}}&\textbf{.806 {\color{imp}\textsubscript{+15.3\%}}}\\

        \bottomrule[1.5pt]
    \end{tabular}
    }
    \label{tab: main_results}
\vspace{-1em}
\end{table}

%% file: tables/conditions.tex
\definecolor{Gray}{gray}{0.9}
\definecolor{imp}{rgb}{0.22, 0.56, 0.24}
\definecolor{dro}{rgb}{0.83, 0.18, 0.18}

\begin{table}[t]
    \centering
\caption{\textbf{Distributed information and unreliable shared context drive the difficulty.}
\textsc{Dist.}, \textsc{Unrel. P}, and \textsc{Unrel. S} denote distributed information, unreliable private evidence, and unreliable shared context, respectively. Distribution alone degrades T2, while shared-context corruption causes the largest additional drop.} 
\vspace{0.3em}
    \resizebox{\linewidth}{!}{
    \begin{tabular}{l|ccc|lll|lll}
        \toprule[1.5pt]
        \multirow{2}{*}{\textbf{Settings}}&\multirow{2}{*}{\textsc{Dist.}}&\multirow{2}{*}{\textsc{Unrel. P}}&\multirow{2}{*}{\textsc{Unrel. S}}&\multicolumn{3}{c|}{\textbf{\textsc{Healthcare-S}}}&\multicolumn{3}{c}{\textbf{\textsc{Healthcare-I}}}\\
        &&&&\textbf{T1}&\textbf{T2}&\textbf{T3}&\textbf{T1}&\textbf{T2}&\textbf{T3}\\
\midrule[1pt]
Oracle&\xmark&\xmark&\xmark&.653&.882&.943&.680&.887&.910\\
Distributed &\cmark&\xmark&\xmark&\textemdash&.580 {\color{dro}\textsubscript{-34.2\%}}&.878 {\color{dro}\textsubscript{-6.9\%}}&\textemdash&.583 {\color{dro}\textsubscript{-34.3\%}}&.797 {\color{dro}\textsubscript{-12.4\%}}\\
Unrel. Shared &\cmark&\xmark&\cmark&\textemdash&.301 {\color{dro}\textsubscript{-65.9\%}}&.708 {\color{dro}\textsubscript{-24.9\%}}&\textemdash&.296 {\color{dro}\textsubscript{-66.6\%}}&.671 {\color{dro}\textsubscript{-26.3\%}}\\
Unrel. Private&\cmark&\cmark&\xmark&.152 {\color{dro}\textsubscript{-76.7\%}}&.546 {\color{dro}\textsubscript{-38.1\%}}&.772 {\color{dro}\textsubscript{-18.1\%}}&.140 {\color{dro}\textsubscript{-79.4\%}}&.560 {\color{dro}\textsubscript{-36.9\%}}&.714 {\color{dro}\textsubscript{-21.5\%}}\\
Standard &\cmark&\cmark&\cmark&.152 {\color{dro}\textsubscript{-76.7\%}}&.273 {\color{dro}\textsubscript{-69.0\%}}&.669 {\color{dro}\textsubscript{-29.1\%}}&.140 {\color{dro}\textsubscript{-79.4\%}}&.287 {\color{dro}\textsubscript{-67.6\%}}&.627 {\color{dro}\textsubscript{-31.1\%}}\\
        \bottomrule[1.5pt]
    \end{tabular}
    }
\vspace{-1em}
    \label{tab: conditions}
\end{table}

%% file: tables/baselines.tex
\begin{wraptable}{R}{0.6\textwidth}
    \centering
\vspace{-1em}
\caption{\textbf{Comparison with representative collaboration strategies.}
D1--D3 denote T3 performance on \textsc{Healthcare-S}, \textsc{Healthcare-I}, and \textsc{Disaster Response}, respectively. \ours{} consistently improves downstream task resolution across both model backbones and all three settings.}
\vspace{0.3em}
    \small
    \begin{tabular}{l|ccc|ccc}
        \toprule[1.5pt]
        \multirow{2}{*}{\textbf{Baselines}}&\multicolumn{3}{c}{\textbf{GPT-4o}}&\multicolumn{3}{c}{\textbf{Gemini-3.5-Flash}}\\
        \cmidrule[0.5pt](lr){2-4}\cmidrule[0.5pt](lr){5-7}
        &D1&D2&D3&D1&D2&D3\\
        \midrule[1pt]
        \textbf{Voting}&.232&.080&.591&.324&.172&.527\\
        \textbf{Robin}&.603&.578&.742&.801&.831&.602\\
        \textbf{Manager}&.658&.562&.602&.871&.865&.591\\
        \textbf{Instructor}&.444&.395&.548&.867&.845&.602\\
        \textbf{Self-Reflect}&.411&.319&.763&.776&.744&.613\\
        \textbf{Confidence}&.526&.341&.699&.858&.843&.591\\
        \midrule[1pt]
        \textbf{Ours}&\textbf{.766}&\textbf{.661}&\textbf{.806}&\textbf{.900}&\textbf{.871}&\textbf{.645}\\
        \bottomrule[1.5pt]
    \end{tabular}
    \label{tab: baseline}
\end{wraptable}

%% file: tables/statistics.tex
\begin{wraptable}{R}{0.6\textwidth}
    \centering
    \small
    \setlength{\tabcolsep}{5pt}
    \renewcommand{\arraystretch}{1.15}
    \vspace{-20pt}
    \caption{Statistics of \ds{}.}
    \label{tab: dataset_statistics}
    \begin{tabular}{lrrrr}
        \toprule[1.5pt]
        \multirow{2}{*}{\textbf{Dataset}}&
        \multirow{2}{*}{\textbf{\# Inst.}} &
        \multicolumn{2}{c}{\textbf{Context}} &
        \multirow{2}{*}{\textbf{\# Agents}} \\
        \cmidrule(lr){3-4}
        & & \textbf{Words.} & \textbf{Sents.} & \\
        \midrule[1pt]
        \textsc{Healthcare}-S & 723&94.19&6.48&7.44\\
        \textsc{Healthcare}-I & 587&94.49&6.54&7.55\\
        \textsc{Disaster Response}  & 93&148.03&6.67&5.55\\
        \bottomrule[1.5pt]
    \end{tabular}
\end{wraptable}

%% file: tables/llm_as_a_judge.tex
\begin{table}[t]
\centering
\caption{\textbf{Consistency across different LLM judges.} Results are averaged over \textsc{Healthcare} S and \textsc{Healthcare} I. Despite differences in absolute scores, GPT-4o, Gemini-3.5-Flash, and Qwen3-235B-A22B induce identical rankings over all seven evaluation settings, yielding pairwise Spearman rank correlations of 1.00.}
\small
\setlength{\tabcolsep}{5pt}
\begin{tabular}{l|ccccccc}
\toprule[1.5pt]
\multirow{2}{*}{\textbf{Judges}}&\multirow{2}{*}{\textbf{Oracle}}&\multicolumn{2}{c}{\textbf{GPT-4o}}&\multicolumn{2}{c}{\textbf{Gemini-3.5-Flash}}&\multicolumn{2}{c}{\textbf{Qwen3-32B}}\\
\cmidrule[0.5pt](lr){3-4}\cmidrule[0.5pt](lr){5-6}\cmidrule[0.5pt](lr){7-8}
&&Stand.&Ours&Stand.&Ours&Stand.&Ours\\
\midrule[1pt]
GPT-4o&.885&.280&.730&.443&.816&.424&.658\\
Gemini-3.5-Flash&.884&.223&.708&.404&.803&.380&.626\\
Qwen3-235B-A22B&.880&.219&.725&.406&.809&.380&.634\\
\bottomrule[1.5pt]
\end{tabular}
\label{tab: llm_as_a_judge}
\end{table}

%% file: tables/mae.tex
\begin{table}[t]
\centering
\caption{\textbf{Fine-grained consistency across different LLM judges.} We report the mean absolute error (MAE) between the scores assigned by each pair of judges to the same individual instances under each evaluation setting. Lower values indicate stronger inter-judge agreement. The consistently small MAEs across judge pairs and evaluation settings demonstrate stable instance-level scoring on both \textsc{Healthcare} S and \textsc{Healthcare} I.}
\small
\setlength{\tabcolsep}{5pt}
\begin{tabular}{l|ccccccc|c}
\toprule[1.5pt]
\multirow{2}{*}{\textbf{Pairs}}&\multirow{2}{*}{\textbf{Oracle}}&\multicolumn{2}{c}{\textbf{GPT-4o}}&\multicolumn{2}{c}{\textbf{Gemini-3.5-Flash}}&\multicolumn{2}{c}{\textbf{Qwen3-32B}}&\multirow{2}{*}{\textbf{Avg.}}\\
\cmidrule[0.5pt](lr){3-4}\cmidrule[0.5pt](lr){5-6}\cmidrule[0.5pt](lr){7-8}
&&Stand.&Ours&Stand.&Ours&Stand.&Ours\\
\midrule[1pt]
\multicolumn{9}{c}{\textsc{Healthcare} S}\\
\midrule[1pt]
GPT vs. Gemini&.096&.146&.141&.167&.126&.164&.152&.142\\
GPT vs. Qwen&.093&.143&.130&.159&.116&.156&.132&.133\\
Gemini vs. Qwen&.091&.114&.134&.144&.123&.137&.147&.127\\
\midrule[1pt]
\multicolumn{9}{c}{\textsc{Healthcare} I}\\
\midrule[1pt]
GPT vs. Gemini&.095&.136&.138&.153&.125&.153&.146&.135\\
GPT vs. Qwen&.091&.134&.132&.145&.118&.150&.146&.131\\
Gemini vs. Qwen&.082&.121&.129&.142&.117&.141&.143&.125\\
\bottomrule[1.5pt]
\end{tabular}
\label{tab: mae}
\end{table}

%% file: tables/human.tex
\begin{wraptable}{R}{0.5\textwidth}
    \centering
    \small
    \setlength{\tabcolsep}{5pt}
    \renewcommand{\arraystretch}{1.15}
    \caption{\textbf{Human evaluation of the GPT-4o fact-level judge.} We report raw agreement and Cohen’s $\kappa$ between two human annotators and GPT-4o on context–fact verification. GPT-4o achieves substantial agreement with both human annotators, supporting its use as an automatic judge for T2 evaluation.}
    \label{tab: human}
    \begin{tabular}{lcc}
        \toprule[1.5pt]
        Settings&Agreement&Cohen's $\kappa$\\
        \midrule[1pt]
        Human 1 vs. Human 2&0.92&0.84\\
        Human 1 vs. GPT-4o&0.88&0.76\\
        Human 2 vs. GPT-4o&0.84&0.68\\
        \bottomrule[1.5pt]
    \end{tabular}
\end{wraptable}

%% file: tables/model_cards.tex
\begin{table}[t]
\centering
\caption{\textbf{Inference configuration.} Unless otherwise specified, we use a temperature of 0 (if available) and keep other settings default.}
\resizebox{\linewidth}{!}{
\begin{tabular}{l|llll}
\toprule[1.5pt]
Model&Provider&Model Identifier&Parameters&Access\\
\midrule[1pt]
Qwen3-8B~\citep{yang2025qwen3}&Alibaba&qwen3-8b&8B&API\\
Qwen3-32B~\citep{yang2025qwen3}&Alibaba&qwen3-32b&32B&API\\
Qwen3-235B-A22B~\citep{yang2025qwen3}&Alibaba&qwen3-235b-a22b&235B / 22B active&API\\
GPT-4o~\citep{hurst2024gpt}&OpenAI&gpt-4o&Not disclosed&API\\
GPT-5-Mini~\citep{singh2025openai}&OpenAI&gpt-5-mini&Not disclosed&API\\
GPT-5~\citep{singh2025openai}&OpenAI&gpt-5&Not disclosed&API\\
Gemini-3.5-Flash~\citep{googledeepmind2026gemini35flash}&Google&gemini-3.5-flash&Not disclosed&API\\
\bottomrule[1.5pt]
\end{tabular}
}
\label{tab: cards}
\end{table}

%% file: tables/main_reliable.tex
\definecolor{Gray}{gray}{0.9}
\definecolor{imp}{rgb}{0.22, 0.56, 0.24}
\definecolor{dro}{rgb}{0.83, 0.18, 0.18}

\begin{table}[t]
    \centering
    \caption{The performance of multi-agent systems powered by different large language models under different discussion strategies, facing the shared information reliability setting.}
    
    \resizebox{\linewidth}{!}{
    \begin{tabular}{l|lll|lll|lll}
        \toprule[1.5pt]
        \multirow{2}{*}{\textbf{Settings}}&\multicolumn{3}{c|}{\textbf{\textsc{Healthcare} S}}&\multicolumn{3}{c|}{\textbf{\textsc{Healthcare} I}}&\multicolumn{3}{c}{\textbf{\textsc{Disaster Response}}}\\
        \cmidrule[0.5pt](lr){2-4}\cmidrule[0.5pt](lr){5-7}\cmidrule[0.5pt](lr){8-10}
        &\textbf{T1}&\textbf{T2}&\textbf{T3}&\textbf{T1}&\textbf{T2}&\textbf{T3}&\textbf{T1}&\textbf{T2}&\textbf{T3}\\
\midrule[1pt]
\multicolumn{10}{l}{\cellcolor{Gray}\textbf{Qwen3-8B}}\\
\midrule[0.5pt]
Oracle&.621&.882&.675&.642&.887&.586&.613&.867&.710\\
Stand.&.140 {\color{dro}\textsubscript{-77.5\%}}&.579 {\color{dro}\textsubscript{-34.4\%}}&.390 {\color{dro}\textsubscript{-42.2\%}}&.155 {\color{dro}\textsubscript{-75.9\%}}&.576 {\color{dro}\textsubscript{-35.1\%}}&.390 {\color{dro}\textsubscript{-33.4\%}}&.129 {\color{dro}\textsubscript{-79.0\%}}&.718 {\color{dro}\textsubscript{-17.2\%}}&.559 {\color{dro}\textsubscript{-21.3\%}}\\
Ours&\textbf{.392 {\color{imp}\textsubscript{+180.0\%}}}&\textbf{.766 {\color{imp}\textsubscript{+32.3\%}}}&\textbf{.505 {\color{imp}\textsubscript{+29.5\%}}}&\textbf{.392 {\color{imp}\textsubscript{+152.9\%}}}&\textbf{.781 {\color{imp}\textsubscript{+35.6\%}}}&\textbf{.465 {\color{imp}\textsubscript{+19.2\%}}}&\textbf{.656 {\color{imp}\textsubscript{+408.5\%}}}&\textbf{.814 {\color{imp}\textsubscript{+13.4\%}}}&\textbf{.645 {\color{imp}\textsubscript{+15.4\%}}}\\
\midrule[1pt]
\multicolumn{10}{l}{\cellcolor{Gray}\textbf{Qwen3-32B}}\\
\midrule[0.5pt]
Oracle&.656&.882&.924&.677&.887&.870&.742&.867&.817\\
Stand.&.126 {\color{dro}\textsubscript{-80.8\%}}&.580 {\color{dro}\textsubscript{-34.2\%}}&.740 {\color{dro}\textsubscript{-19.9\%}}&.090 {\color{dro}\textsubscript{-86.7\%}}&.583 {\color{dro}\textsubscript{-34.3\%}}&.652 {\color{dro}\textsubscript{-25.1\%}}&.172 {\color{dro}\textsubscript{-76.8\%}}&.676 {\color{dro}\textsubscript{-22.0\%}}&.559 {\color{dro}\textsubscript{-31.6\%}}\\
Ours&\textbf{.394 {\color{imp}\textsubscript{+212.7\%}}}&\textbf{.742 {\color{imp}\textsubscript{+27.9\%}}}&\textbf{.783 {\color{imp}\textsubscript{+5.8\%}}}&\textbf{.355 {\color{imp}\textsubscript{+294.4\%}}}&\textbf{.775 {\color{imp}\textsubscript{+32.9\%}}}&\textbf{.771 {\color{imp}\textsubscript{+18.3\%}}}&\textbf{.630 {\color{imp}\textsubscript{+266.3\%}}}&\textbf{.806 {\color{imp}\textsubscript{+19.2\%}}}&\textbf{.685 {\color{imp}\textsubscript{+22.5\%}}}\\
\midrule[1pt]
\multicolumn{10}{l}{\cellcolor{Gray}\textbf{Qwen3-235B-A22B}}\\
\midrule[0.5pt]
Oracle&.671&.882&.920&.659&.887&.865&.742&.867&.849\\
Stand.&.127 {\color{dro}\textsubscript{-81.1\%}}&.569 {\color{dro}\textsubscript{-35.5\%}}&.746 {\color{dro}\textsubscript{-18.9\%}}&.112 {\color{dro}\textsubscript{-83.0\%}}&.569 {\color{dro}\textsubscript{-35.9\%}}&.562 {\color{dro}\textsubscript{-35.0\%}}&.161 {\color{dro}\textsubscript{-78.3\%}}&.687 {\color{dro}\textsubscript{-20.8\%}}&.656 {\color{dro}\textsubscript{-22.7\%}}\\
Ours&\textbf{.426 {\color{imp}\textsubscript{+235.4\%}}}&\textbf{.771 {\color{imp}\textsubscript{+35.5\%}}}&\textbf{.769 {\color{imp}\textsubscript{+3.1\%}}}&\textbf{.390 {\color{imp}\textsubscript{+248.2\%}}}&\textbf{.781 {\color{imp}\textsubscript{+37.3\%}}}&\textbf{.736 {\color{imp}\textsubscript{+31.0\%}}}&\textbf{.559 {\color{imp}\textsubscript{+247.2\%}}}&\textbf{.806 {\color{imp}\textsubscript{+17.3\%}}}&\textbf{.720 {\color{imp}\textsubscript{+9.8\%}}}\\
\midrule[1pt]
\multicolumn{10}{l}{\cellcolor{Gray}\textbf{Gemini-3.5-Flash}}\\
\midrule[0.5pt]
Oracle&.849&.882&.963&.835&.887&.961&.914&.867&.817\\
Stand.&.144 {\color{dro}\textsubscript{-83.0\%}}&.604 {\color{dro}\textsubscript{-31.5\%}}&.870 {\color{dro}\textsubscript{-9.7\%}}&.118 {\color{dro}\textsubscript{-85.9\%}}&.610 {\color{dro}\textsubscript{-31.2\%}}&\textbf{.874 {\color{dro}\textsubscript{-9.1\%}}}&.140 {\color{dro}\textsubscript{-84.7\%}}&.735 {\color{dro}\textsubscript{-15.2\%}}&.570 {\color{dro}\textsubscript{-30.2\%}}\\
Ours&\textbf{.676 {\color{imp}\textsubscript{+369.4\%}}}&\textbf{.828 {\color{imp}\textsubscript{+37.1\%}}}&\textbf{.899 {\color{imp}\textsubscript{+3.3\%}}}&\textbf{.634 {\color{imp}\textsubscript{+437.3\%}}}&\textbf{.823 {\color{imp}\textsubscript{+34.9\%}}}&.869 {\color{dro}\textsubscript{-0.6\%}}&\textbf{.903 {\color{imp}\textsubscript{+545.0\%}}}&\textbf{.855 {\color{imp}\textsubscript{+16.3\%}}}&\textbf{.613 {\color{imp}\textsubscript{+7.5\%}}}\\
\midrule[1pt]
\multicolumn{10}{l}{\cellcolor{Gray}\textbf{GPT-5-Mini}}\\
\midrule[0.5pt]
Oracle&.746&.882&.971&.763&.887&.957&.914&.867&.839\\
Stand.&.163 {\color{dro}\textsubscript{-78.2\%}}&.537 {\color{dro}\textsubscript{-39.1\%}}&.816 {\color{dro}\textsubscript{-16.0\%}}&.143 {\color{dro}\textsubscript{-81.3\%}}&.538 {\color{dro}\textsubscript{-39.3\%}}&.789 {\color{dro}\textsubscript{-17.6\%}}&.172 {\color{dro}\textsubscript{-81.2\%}}&.626 {\color{dro}\textsubscript{-27.8\%}}&.731 {\color{dro}\textsubscript{-12.9\%}}\\
Ours&\textbf{.593 {\color{imp}\textsubscript{+263.8\%}}}&\textbf{.810 {\color{imp}\textsubscript{+50.8\%}}}&\textbf{.845 {\color{imp}\textsubscript{+3.6\%}}}&\textbf{.561 {\color{imp}\textsubscript{+292.3\%}}}&\textbf{.807 {\color{imp}\textsubscript{+50.0\%}}}&\textbf{.804 {\color{imp}\textsubscript{+1.9\%}}}&\textbf{.806 {\color{imp}\textsubscript{+368.6\%}}}&\textbf{.823 {\color{imp}\textsubscript{+31.5\%}}}&\textbf{.806 {\color{imp}\textsubscript{+10.3\%}}}\\
\midrule[1pt]
\multicolumn{10}{l}{\cellcolor{Gray}\textbf{GPT-5}}\\
\midrule[0.5pt]
Oracle&.805&.882&.968&.787&.887&.968&.882&.867&.892\\
Stand.&.109 {\color{dro}\textsubscript{-86.5\%}}&.542 {\color{dro}\textsubscript{-38.5\%}}&.888 {\color{dro}\textsubscript{-8.3\%}}&.124 {\color{dro}\textsubscript{-84.2\%}}&.544 {\color{dro}\textsubscript{-38.7\%}}&.829 {\color{dro}\textsubscript{-14.4\%}}&.183 {\color{dro}\textsubscript{-79.3\%}}&.631 {\color{dro}\textsubscript{-27.2\%}}&.699 {\color{dro}\textsubscript{-21.6\%}}\\
Ours&\textbf{.556 {\color{imp}\textsubscript{+410.1\%}}}&\textbf{.815 {\color{imp}\textsubscript{+50.4\%}}}&\textbf{.902 {\color{imp}\textsubscript{+1.6\%}}}&\textbf{.535 {\color{imp}\textsubscript{+331.5\%}}}&\textbf{.821 {\color{imp}\textsubscript{+50.9\%}}}&\textbf{.860 {\color{imp}\textsubscript{+3.7\%}}}&\textbf{.839 {\color{imp}\textsubscript{+358.5\%}}}&\textbf{.830 {\color{imp}\textsubscript{+31.5\%}}}&\textbf{.849 {\color{imp}\textsubscript{+21.5\%}}}\\
\midrule[1pt]
\multicolumn{10}{l}{\cellcolor{Gray}\textbf{GPT-4o}}\\
\midrule[0.5pt]
Oracle&.653&.882&.943&.680&.887&.910&.817&.867&.925\\
Stand.&.152 {\color{dro}\textsubscript{-76.7\%}}&.546 {\color{dro}\textsubscript{-38.1\%}}&.772 {\color{dro}\textsubscript{-18.1\%}}&.140 {\color{dro}\textsubscript{-79.4\%}}&.560 {\color{dro}\textsubscript{-36.9\%}}&\textbf{.714 {\color{dro}\textsubscript{-21.5\%}}}&.194 {\color{dro}\textsubscript{-76.3\%}}&.751 {\color{dro}\textsubscript{-13.4\%}}&.731 {\color{dro}\textsubscript{-21.0\%}}\\
Robin&.154 {\color{dro}\textsubscript{-76.4\%}}&.578 {\color{dro}\textsubscript{-34.5\%}}&.770 {\color{dro}\textsubscript{-18.3\%}}&.167 {\color{dro}\textsubscript{-75.4\%}}&.588 {\color{dro}\textsubscript{-33.7\%}}&.685 {\color{dro}\textsubscript{-24.7\%}}&.172 {\color{dro}\textsubscript{-78.9\%}}&.774 {\color{dro}\textsubscript{-10.7\%}}&.774 {\color{dro}\textsubscript{-16.3\%}}\\
Chain&.158 {\color{dro}\textsubscript{-75.8\%}}&.575 {\color{dro}\textsubscript{-34.8\%}}&.780 {\color{dro}\textsubscript{-17.3\%}}&.187 {\color{dro}\textsubscript{-72.5\%}}&.586 {\color{dro}\textsubscript{-33.9\%}}&.705 {\color{dro}\textsubscript{-22.5\%}}&.226 {\color{dro}\textsubscript{-72.3\%}}&.763 {\color{dro}\textsubscript{-12.0\%}}&.753 {\color{dro}\textsubscript{-18.6\%}}\\
Tree&.136 {\color{dro}\textsubscript{-79.2\%}}&.561 {\color{dro}\textsubscript{-36.4\%}}&.747 {\color{dro}\textsubscript{-20.8\%}}&.175 {\color{dro}\textsubscript{-74.3\%}}&.572 {\color{dro}\textsubscript{-35.5\%}}&.678 {\color{dro}\textsubscript{-25.5\%}}&.129 {\color{dro}\textsubscript{-84.2\%}}&.760 {\color{dro}\textsubscript{-12.3\%}}&.742 {\color{dro}\textsubscript{-19.8\%}}\\
Ours&\textbf{.567 {\color{imp}\textsubscript{+273.0\%}}}&\textbf{.793 {\color{imp}\textsubscript{+45.2\%}}}&\textbf{.823 {\color{imp}\textsubscript{+6.6\%}}}&\textbf{.545 {\color{imp}\textsubscript{+289.3\%}}}&\textbf{.797 {\color{imp}\textsubscript{+42.3\%}}}&.702 {\color{dro}\textsubscript{-1.7\%}}&\textbf{.785 {\color{imp}\textsubscript{+304.6\%}}}&\textbf{.845 {\color{imp}\textsubscript{+12.5\%}}}&\textbf{.806 {\color{imp}\textsubscript{+10.3\%}}}\\

        \bottomrule[1.5pt]
    \end{tabular}
    }
    \label{tab: main_reliable}
\end{table}

%% file: tables/ablation.tex
\begin{wraptable}{R}{0.45\textwidth}
    \centering
    \caption{\textbf{Ablation on the candidate pool size $M$.} $M$ determines both the number of candidate-selection iterations and the resulting candidate pool size. We vary $M$ from 1 to 5 and additionally report an Oracle setting in which the misleading agent is directly identified. Smaller or overly large candidate pools generally degrade performance, while the Oracle results highlight the impact of source-attribution errors on subsequent belief repair and decision making.}
    \resizebox{\linewidth}{!}{
    \begin{tabular}{l|ccc}
        \toprule[1.5pt]
        Settings&T1&T2&T3\\
        \midrule[1pt]
        Vanilla&.545&.728&.661\\
        \midrule[1pt]
        $M=1$&.413 {\color{dro}\textsubscript{-24.2\%}}&.645 {\color{dro}\textsubscript{-11.4\%}}&.611 {\color{dro}\textsubscript{-7.6\%}}\\
        $M=2$&.452 {\color{dro}\textsubscript{-17.1\%}}&.696 {\color{dro}\textsubscript{-4.4\%}}&.654 {\color{dro}\textsubscript{-1.1\%}}\\
        $M=4$&.558 {\color{imp}\textsubscript{+2.4\%}}&.736 {\color{imp}\textsubscript{+1.1\%}}&.671 {\color{imp}\textsubscript{+1.5\%}}\\
        $M=5$&.496 {\color{dro}\textsubscript{-9.0\%}}&.710 {\color{dro}\textsubscript{-2.5\%}}&.638 {\color{dro}\textsubscript{-3.5\%}}\\
        Oracle&1.00 {\color{imp}\textsubscript{+83.5\%}}&.813 {\color{imp}\textsubscript{+11.7\%}}&.837 {\color{imp}\textsubscript{+26.6\%}}\\
        \bottomrule[1.5pt]
    \end{tabular}
    }
    \label{tab: ablation}
\end{wraptable}

%% file: tables/token_cost.tex
\begin{table}[t]
\centering
\caption{\textbf{Token consumption of different collaboration frameworks.} We report the average number of input, output, and total tokens required per instance for GPT-4o and Gemini-3.5-Flash across three datasets. All values are reported in thousands of tokens ($\times 10^3$). \ours{} maintains moderate token consumption compared with discussion-intensive collaboration baselines.}
\small
\setlength{\tabcolsep}{5pt}
\begin{tabular}{l|ccc|ccc|ccc}
\toprule[1.5pt]
\multirow{2}{*}{\textbf{Settings}}&\multicolumn{3}{c}{\textbf{\textsc{Healthcare}-S}}&\multicolumn{3}{c}{\textbf{\textsc{Healthcare}-I}}&\multicolumn{3}{c}{\textbf{\textsc{Disaster Response}}}\\
\cmidrule[0.5pt](lr){2-4}\cmidrule[0.5pt](lr){5-7}\cmidrule[0.5pt](lr){8-10}
&\# Input&\# Output&\# Total&\# Input&\# Output&\# Total&\# Input&\# Output&\# Total\\
\midrule[1pt]
\multicolumn{10}{c}{GPT-4o}\\
\midrule[0.5pt]
Standard&46.3&5.7&52.1&63.6&7.8&71.5&26.9&4.4&31.2\\
Chain&19.9&5.2&25.0&23.3&6.2&29.5&13.3&3.5&16.8\\
Tree&20.2&5.2&25.4&24.0&6.4&30.3&13.6&3.6&17.2\\
Voting&3.5&1.5&5.0&3.6&1.5&5.1&3.6&1.4&5.0\\
Robin&14.8&4.6&19.3&16.5&5.3&21.7&10.2&3.1&13.3\\
Manager&9.0&6.2&15.2&9.0&6.2&15.2&8.3&5.6&13.9\\
Instruction&9.6&0.7&10.4&10.0&0.8&10.8&10.9&0.9&11.8\\
Self-Reflect&66.3&5.2&71.5&76.1&5.8&81.9&37.2&3.4&40.6\\
Confidence&37.9&4.8&42.7&42.6&5.4&48.0&22.4&3.8&26.2\\
\midrule[0.5pt]
Ours&16.2&3.2&19.4&16.8&3.4&20.2&14.7&2.9&17.6\\

\midrule[1pt]
\multicolumn{10}{c}{Gemini-3.5-Flash}\\
\midrule[0.5pt]
Standard&117.2&14.0&131.2&124.5&14.7&139.2&30.9&5.0&35.9\\
Voting&3.7&1.6&5.3&3.7&1.7&5.4&3.7&1.4&5.1\\
Robin&41.6&14.4&56.0&43.3&15.0&58.2&14.9&4.8&19.7\\
Manager&8.9&5.9&14.8&9.2&6.2&15.4&8.2&5.4&13.6\\
Instruction&9.8&0.7&10.4&9.9&0.7&10.6&10.7&0.7&11.4\\
Self-Reflect&80.7&6.6&87.3&76.0&6.0&82.0&41.4&3.8&45.2\\
Confidence&48.0&6.2&54.2&48.7&6.2&54.8&26.1&4.3&30.4\\
\midrule[0.5pt]
Ours&18.9&3.7&22.6&19.5&3.8&23.3&16.5&3.2&19.7\\
\bottomrule[1.5pt]
\end{tabular}
\label{tab: token_cost}
\end{table}